\documentclass[times, review, 10pt]{elsarticle}
\usepackage{xcolor}

\usepackage{adjustbox}
\usepackage{subfig}
\usepackage{float}
\usepackage{svg}
\usepackage{makecell}
\usepackage{amsmath}
\usepackage{amssymb}
\usepackage{amsfonts}
\usepackage{algorithmic}
\usepackage{graphicx}
\usepackage{textcomp}
\usepackage{xcolor}
\usepackage{lscape}
\usepackage{afterpage}
\usepackage{multirow}
\journal{Pattern Recognition}

\begin{document}

\begin{frontmatter}

%% Title, authors and addresses

%% use the tnoteref command within \title for footnotes;
%% use the tnotetext command for theassociated footnote;
%% use the fnref command within \author or \affiliation for footnotes;
%% use the fntext command for theassociated footnote;
%% use the corref command within \author for corresponding author footnotes;
%% use the cortext command for theassociated footnote;
%% use the ead command for the email address,
%% and the form \ead[url] for the home page:
%% \title{Title\tnoteref{label1}}
%% \tnotetext[label1]{}
%% \author{Name\corref{cor1}\fnref{label2}}
%% \ead{email address}
%% \ead[url]{home page}
%% \fntext[label2]{}
%% \cortext[cor1]{}
%% \affiliation{organization={},
%%             addressline={},
%%             city={},
%%             postcode={},
%%             state={},
%%             country={}}
%% \fntext[label3]{}

\title{Continual Graph Learning: A Survey}

%% use optional labels to link authors explicitly to addresses:
%% \author[label1,label2]{}
%% \affiliation[label1]{organization={},
%%             addressline={},
%%             city={},
%%             postcode={},
%%             state={},
%%             country={}}
%%
%% \affiliation[label2]{organization={},
%%             addressline={},
%%             city={},
%%             postcode={},
%%             state={},
%%             country={}}

\author[1,2]{QiAo Yuan}

% Corresponding author indication
% \cormark[1]

% Footnote of the first author
% \fnmark[1]

% Email id of the first author
\ead{qiao.yuan17@student.xjtlu.edu.cn}

% % URL of the first author
% \ead[url]{www.cvr.cc, cvr@sayahna.org}

% Address/affiliation
\affiliation[1]{organization={University of Liverpool},
    % addressline={Radarweg 29}, 
    city={Liverpool},
    % citysep={}, % Uncomment if no comma needed between city and postcode
    postcode={L69 3BX}, 
    % state={},
    country={United Kindom}}

% Second author
\author[2]{Sheng-Uei Guan \corref{cor1}}

\cortext[cor1]{Corresponding author}
 
% \fnmark[2]
\ead{Steven.Guan@xjtlu.edu.cn}

% Third author
\author[3]{Pin Ni}
% \fnmark[2]
\ead{pin.ni.21@ucl.ac.uk}

% Address/affiliation
\affiliation[2]{organization={Xi’an Jiaotong-Liverpool University},
    % addressline={}, 
    city={Suzhou},
    % citysep={}, % Uncomment if no comma needed between city and postcode
    postcode={215123}, 
    % state={Trivandrum},
    country={China}}

% Fourth author
\author%
[1,2]
{Tianlun Luo}
% \cormark[2]
% \fnmark[1,3]
\ead{tianlun.luo@liverpool.ac.uk}
\

\affiliation[3]{organization={University College London},
    % addressline={Mepukada}, 
    city={London},
    % citysep={}, % Uncomment if no comma needed between city and postcode
    postcode={WC1E 6BT}, 
    % state={Trivandrum},
    country={United Kindom}}

\author%
[2]
{Ka Lok Man}
% \cormark[2]
% \fnmark[1,3]
\ead{Ka.Man@xjtlu.edu.cn}

\author%
[1]
{Prudence Wong}
% \cormark[2]
% \fnmark[1,3]
\ead{pwong@liverpool.ac.uk}

\author%
[4]
{Victor Chang}

% \cormark[2]
% \fnmark[1,3]
\ead{victorchang.research@gmail.com}

\affiliation[4]{organization={Aston University},
    % addressline={Mepukada}, 
    city={Birmingham},
    % citysep={}, % Uncomment if no comma needed between city and postcode
    postcode={B4 7ET}, 
    % state={Trivandrum},
    country={United Kindom}}

%% Abstract
\begin{abstract}
Research in continual learning (CL) has primarily focused on data represented in Euclidean space, with comparatively limited attention given to graph-structured data. Moreover, most existing graph learning models are designed for static graphs, yet in real-world applications, graphs typically evolve over time. This dynamic nature of graphs introduces challenges such as catastrophic forgetting, which also affects graph learning models when trained incrementally. As a result, there is an urgent need to develop robust, effective, and efficient methods for continual graph learning (CGL). CGL is an emerging field that aims to extend continual learning to graph-structured data. This paper introduces the core concepts of CGL, distinguishing it from related areas, and provides a comprehensive review of recent studies. These studies are categorized based on the strategies employed to address the unique challenges posed by graph-structured data. The paper also highlights key challenges in CGL, supported by experimental analysis, and offers potential solutions to these challenges. Finally, it presents a curated collection of datasets, code repositories, and benchmarks, and outlines open issues and future research directions in this rapidly evolving field.
\end{abstract}

\begin{highlights}
\item Conducts in-depth analysis of CGL literature with extensive empirical comparisons.
\item Benchmarks effectiveness and efficiency of ten representative methods across five datasets, two GNN backbones, and two continual learning settings.
\item Lists representative datasets and open-source implementations for reproducibility.
\item Proposes a practical roadmap for developing continual graph learning methods.
\end{highlights}

%% Keywords

\end{frontmatter}

%% Add \usepackage{lineno} before \begin{document} and uncomment 
%% following line to enable line numbers
%% \linenumbers

%% main text
%%

%% Use \section commands to start a section
\section{Introduction}

Continual learning (CL), or lifelong learning, aims to incrementally acquire knowledge from sequential data streams while retaining prior information to facilitate future learning tasks \cite{CGL_survey2019, CGL_survey2021}. Although neural networks achieve human-level performance on isolated tasks under controlled conditions, their efficacy diminishes significantly when processing continuous data streams due to catastrophic forgetting (CF) \cite{MCCLOSKEY1989109}. The impracticality of full retraining, due to data privacy concerns, storage limitations, and computational costs, necessitates alternative strategies for mitigating catastrophic forgetting (CF) \cite{CGL_survey2021}. Effective CL requires balancing plasticity (adapting to new knowledge) and stability (retaining existing knowledge), a fundamental challenge known as the stability-plasticity dilemma \cite{SP-delimma}. While CL research predominantly focuses on Euclidean data, its application to graph-structured data remains underexplored.

Graphs provide a versatile representation of relational data through node-edge structures, enabling applications in citation networks, social systems, and traffic modeling \cite{gl_survey}. Graph learning (GL) integrates structural and feature-based relationships into component representations for downstream tasks. Real-world graph dynamics motivate dynamic graph learning (DGL) approaches that incrementally update models while preserving historical patterns \cite{DGL_survey}. However, current DGL research emphasizes transfer learning over CF mitigation in graph scenarios, as evidenced by limited attention in works like \cite{yao2024recurrent} and \cite{chen2025signn}.

Continual graph learning (CGL) is an emerging field focused on addressing catastrophic forgetting (CF) in evolving graph environments. Unlike conventional continual learning (CL), CGL introduces two unique challenges: node-level dependencies (interconnected nodes) and task-level dependencies (relationships between past and new graph structures). Resolving these dependencies is crucial for achieving the stability-plasticity balance in graph-based continual learning. However, existing surveys primarily offer descriptive taxonomies, leaving a critical gap in systematically analyzing the operational strengths and weaknesses of these methods under unified experimental protocols. To address this, this survey moves beyond descriptive taxonomies to conduct a critical analysis of the operational strengths and weaknesses of existing CGL methodologies. We substantiate this analysis through unified empirical benchmarking, exposing performance trade-offs often obscured in prior literature. Furthermore, to bridge the gap between theory and practice, we provide curated datasets, code repositories, and discussions for future research direction.

\section{Background}

\subsection{Continual Learning}

Continual learning (CL) addresses dynamic data distributions by incrementally learning from a sequence of tasks. A key challenge is catastrophic forgetting, where performance on previous tasks deteriorates as new tasks are learned. CL seeks to balance stability (preserving knowledge from earlier tasks) and plasticity (adapting to new tasks). Formally, CL involves solving a sequence of sub-tasks $\mathcal{T} = \{t\}_{t=1}^T$, with the goal of optimizing performance across all tasks. Each task $t$ has a dataset $D^t$, and the total loss over the $T$ tasks is: $\mathcal{L}_{1:T}(\theta) = \mathcal{L}_{1:T-1}(\theta) + \mathcal{L}_T(\theta)$ where $\mathcal{L}_{1:T-1}(\theta)$ is the loss on previous tasks and $\mathcal{L}_T(\theta)$ is the loss on the current task.

In CL, access to data from previous tasks $D^{1:T-1}$ is typically restricted or unavailable due to resource or privacy constraints, making direct optimization of $\mathcal{L}_{1:T}(\theta)$ intractable. CL algorithms approximate $\mathcal{L}_{1:T-1}(\theta)$ using limited memory, regularization, or architecture-based strategies, enabling efficient updates that approximate the performance of models trained on all data.

\subsection{Related Topics}
Continual Graph Learning (CGL) shares similarities with dynamic graph learning (DGL) and graph domain adaptation (GDA). DGL addresses graph learning problems where the graph evolves, including topological changes, feature updates, and diffusion processes \cite{barros2021survey}. However, DGL differs from CGL in two ways: (1) all data remain accessible throughout the graph's evolution, and (2) it does not prioritize maintaining performance on earlier graph snapshots, focusing instead on capturing evolutionary patterns to address the current task \cite{hedegaard2023continual}. GDA, a subfield of transfer learning, adapts models trained on a source domain to a target domain with different data distributions \cite{liu2023structural}. While both DGL and GDA emphasize forward knowledge transfer, they do not restrict access to previous data. In contrast, CGL aims to balance forward knowledge transfer with knowledge retention, where prior data are inaccessible.

\subsection{Related Works}
Several surveys are partially related to this paper. Febrinanto et al. \cite{febrinanto2023graph} introduce a foundational taxonomy but focus primarily on literature published before 2023 and do not include empirical benchmarking. Tian et al. \cite{tian2024continual} categorize methodologies aimed at performance improvement, yet their analysis remains qualitative and does not address operational advantages and disadvantages in details. Zhang et al. \cite{zhang2024continual} examine continual graph learning (CGL) from a mathematical perspective but do not provide practical implementation resources. Wu et al. \cite{wu2024graph} offer a broad overview of distribution shifts, but this wide scope restricts the depth of discussion regarding CGL-specific implementation. In contrast, our survey extends beyond descriptive summarization by offering a critical analysis of the strengths and weaknesses of current CGL methodologies. We further support this theoretical assessment with comprehensive empirical benchmarking of ten representative methods across five datasets and two continual learning settings. This approach enables us to discuss limitations from a theoretical perspective and to empirically validate performance trade-offs, such as the observed degradation of regularization methods in Class-IL, thereby providing robust evidence to inform future research.

\section{Preliminaries}

\subsection{Problem Statement}
CGL incrementally solves a sequence of sub-tasks $\{T_k\}^K_{k=1}$ with an aim to solve the complete task when all sub-tasks are learned eventually. Each sub-task holds a graph-structured data $(\mathcal{G}^{T_k} = (X^{T_k}, V^{T_k}, E^{T_k}), Y^{T_k})$ where $X^{T_k}$ is the feature, $V^{T_k}$ is the node set, $E^{T_k}$ is the edge set, $Y^{T_k}$ is the label set, and $T_k$ is the task identifier. In most CGL problems, graph evolvement indicates incremental topology update, i.e., $\mathcal{G}^{T_{1:k}} = \mathcal{G}^{T_{1:k-1}} \cup \mathcal{G}^{T_{k}}$, where $\mathcal{G}^{T_{1:k}}$ is the whole graph up to task $T_k$. The node or edge feature is static while only graph topology changes. The dataset of each task is split into a training set $\mathcal{G}^{T_k}_{\operatorname{tr}}$ and a testing set $\mathcal{G}^{T_k}_{\operatorname{te}}$. At task $T_k$, the previous training set $\mathcal{G}^{T_{1:k-1}}$ is inaccessible, the model aims to minimize the loss over all observed testing sets as: $\mathcal{L} = \sum^{K}_{k=1} f(\mathcal{G}^{T_k}_{\operatorname{te}}, Y^{T_k}_{\operatorname{te}})$ where $K$ is the task number. 

\subsection{Graph Learning Setting}

Transductive and inductive learning are key paradigms in graph learning. In inductive learning, a model is trained in a training set and applied to a separate testing set, which may belong to a different graph and is not accessed during training. In contrast, transductive learning involves a testing set that resides within the same graph as the training set, with both data and labels used during inference to predict test labels. Inductive learning is typically more challenging due to the need for generalization to unseen nodes or graphs.

\subsection{Continual Learning Setting}
Based on the accessibility of task identifiers and the properties of output spaces, the continual learning settings in CGL can be categorized into task-incremental (Task-IL), domain-incremental (Domain-IL), and class-incremental (Class-IL) problems \cite{wang2024comprehensive}. In all these settings, the input distributions of different tasks differ. In the Task-IL setting, task identifiers are available during both training and testing. The output spaces of different tasks are disjoint, i.e., $Y^{T_i} \cap Y^{T_j} = \emptyset$ for $i \neq j$. Regarding to the Domain-IL setting, task identifiers are also accessible during both training and testing. However, all tasks share the same output space, i.e., $Y^{T_i} = Y^{T_j}$ for $i \neq j$. In terms of the Class-IL setting, task identifiers are accessible during training but unavailable during testing. Additionally, the output spaces of earlier tasks are subsumed by those of later tasks, i.e., $Y^{T_i} \subset Y^{T_j}$ for $i < j$.

Each setting presents distinct challenges. In Task-IL, the model knows which task it is solving at test time, and separate task-specific output heads can be used, making it easier to avoid confusion between tasks (though catastrophic forgetting can still occur if representations are overwritten). In Domain-IL, the output labels remain the same across tasks, but the input distribution shifts per task. The challenge is handling these domain shifts while using a shared classifier, methods must prevent forgetting earlier domains and also deal with potential domain shift without task cues (aside from knowing the task id to route to the right head, which here is the same head). In Class-IL, which is generally the most challenging scenario, the model is not informed of the task identity at test time and must classify among all classes seen so far. The output space is cumulative and expands with each task. This makes catastrophic forgetting especially acute in Class-IL.

\subsection{Evaluation Metrics}
\label{Eva_me}

Similar to CL, the performance of a well-trained model after observing $T$ tasks is evaluated using four criteria: average performance (AP), forward transfer (FT), average forgetting (AF), and intransigence (INT). Following \cite{GEM}, we define the performance on task $i$ after observing $j$ tasks as $P_{i,j}$ and the performance with randomly initialized parameters as $R_i$. AP measures overall performance across all tasks, FT evaluates the model’s ability to transfer knowledge to future tasks, AF quantifies resistance to catastrophic forgetting by measuring performance degradation on previous tasks, and INT \cite{chaudhry2019tiny} compares joint training performance to CL after $T$ tasks. These metrics are defined as:
$$ AP = \frac{\sum^{T}_{i=1} P_{i,T}}{T}, FT = \frac{\sum^T_{i=2}(P_{i,i-1} - R_i)}{T-1}, AF = \frac{\sum^{T-1}_{i=1}(P_{i,T} - P_{i,i})}{T-1}, INT = \frac{\sum^T_{i=1}(P^{joint}_{i,i}-P_{i,i})}{T}$$
Where $P^{joint}_{i,i}$ denotes the performance of a joint training model on task $i$ after learning $i$ tasks.

\begin{figure*}[htbp!]
\centering
\includegraphics[width=\textwidth]{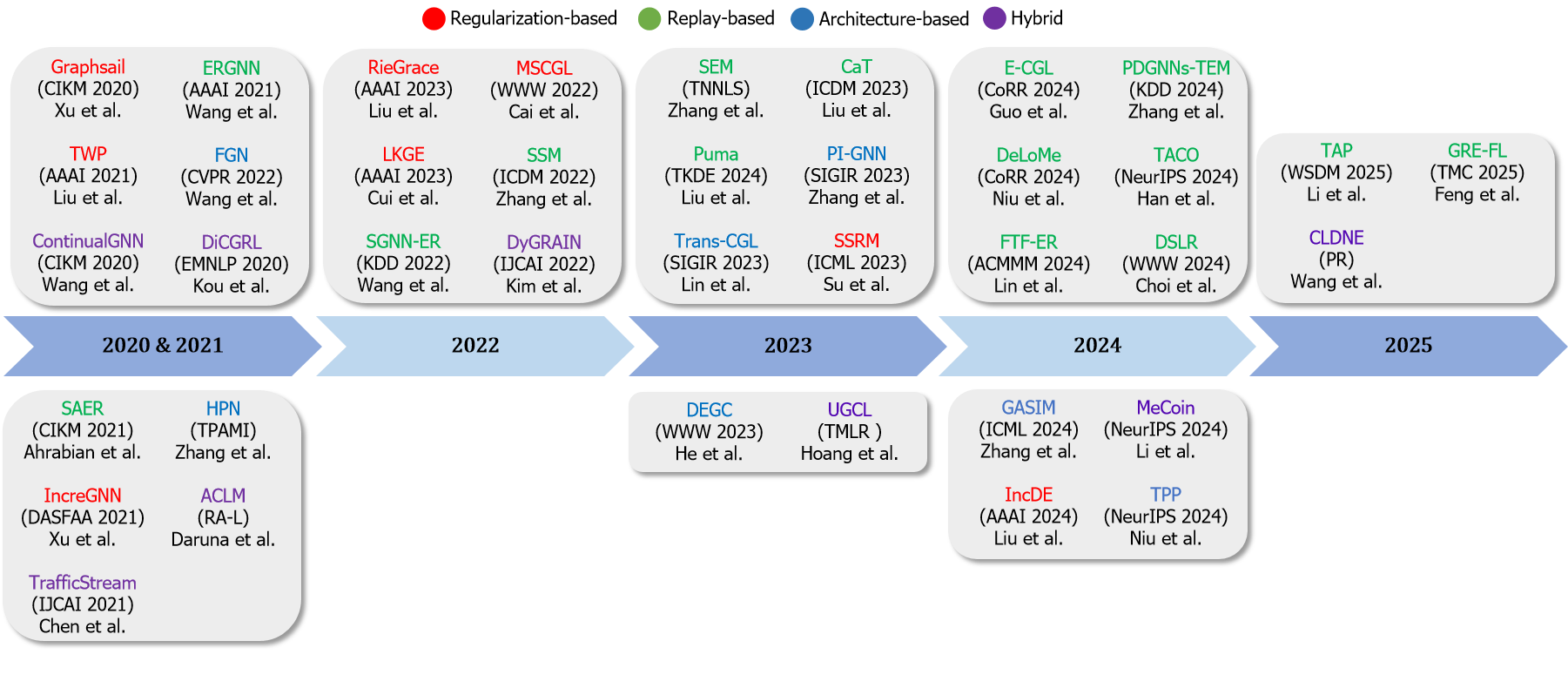}
\caption{Chronological overview of CGL methods from 2020 to 2025. The figure categorizes representative works by publication year and method family, using color codes for visual grouping: red for regularization-based methods, green for replay-based methods, blue for architecture-based methods, and purple for methods with multiple family of CGL strategies.}
\label{timeline}
\end{figure*}

\begin{table}[htp]
\centering
\caption{Summary of Continual Graph Learning (CGL) methods, including their continual learning (CL) setting, graph learning setting, innovation, and application domain. TIL, DIL, and CIL indicate task-incremental, domain-incremental, and class-incremental setting.}
\scriptsize
\begin{adjustbox}{max height=\textheight, max width = \textwidth}
\begin{tabular}{l|c|c|p{5.7cm}|l}
\hline
\textbf{Method} & \textbf{CL Setting} & \textbf{Graph Setting} & \textbf{Innovation} & \textbf{Application} \\
\hline
GraphSAIL \cite{GraphSail} & DIL & Inductive & Distills local/global node positions & Recommendation \\
CGNN \cite{CGNN} & TIL & Transductive & Selects boundary and influenced nodes for replay & Node Classification \\
FGN \cite{FGN} & CIL & Inductive & Reformulates node classification as graph classification & \makecell[l]{Node Classification \\ Action Recognition} \\
DiCGRL \cite{DiCGRL} & TIL, DIL & Inductive & Semantic-specific embedding decomposition & \makecell[l]{Node Classification \\ Link Prediction} \\
TWP \cite{TWP} & TIL & Transductive & Regularizes neighbor interactions & \makecell[l]{Node Classification \\ Graph Classification} \\
ER-GNN \cite{ERGNN} & TIL & Transductive & Selects nodes by cluster mean and coverage maximization & Node Classification \\
SAER \cite{SAER} & DIL & Inductive & Replays low-degree nodes to address long-tail & Recommendation \\
TrafficStream \cite{TrafficStream} & DIL & Inductive & Replays nodes with high distributional drift & Traffic Forecasting \\
IncreGNN \cite{IncreGNN} & TIL & Inductive & Replays affected and high-degree nodes & Recommendation \\
HPN \cite{HPN} & CIL & Unknown & Stores prototypes hierarchically across tasks & Node Classification \\
SGNN-GR \cite{SGNN-GR} & TIL & Unknown & GRU-based walk generation for replay & Node Classification \\
MSCGL \cite{MSCGL} & TIL & Transductive & Applies GNN architecture search and parameter isolation & Node Classification \\
SSM \cite{SSM} & CIL & Transductive & Degree/random sampling for subgraph replay & Node Classification \\
DyGRAIN \cite{DyGRAIN} & Unknown & Unknown & \makecell[l]{Replays impacted nodes \\ Distills severely degraded nodes} & Node Classification \\
DEGC \cite{DEGC} & DIL & Inductive & Parameter isolation for long/short-term interests & Recommendation \\
RieGrace \cite{RieGrace}  & TIL & Transductive & Distills knowledge in hyperbolic space & Node Classification \\
LKGE \cite{LKGE} & DIL & Inductive & Elastic regularization based on past connections & Link Prediction \\
Trans-CGL \cite{trans-CGL} & Unknown & Unknown & \makecell[l]{Gated attention via parameter isolation \\ Asymmetric attention regularization} & Node Classification \\
CaT \cite{CaT} & TIL, CIL & Transductive & Graph condensation for generative replay & Node Classification \\
PI-GNN \cite{PI-GNN} & TIL & Transductive & Parameter isolation for stable/unstable nodes & Node Classification \\
SSRM \cite{SSRM} & TIL & Inductive & Minimizes inter-graph structural shift & Node Classification \\
SEM \cite{SEM} & CIL & Transductive & Samples subgraphs via curvature estimation & Node Classification \\
TACO \cite{TACO} & CIL & Inductive & Coarsens graphs based on representation proximity & Node Classification \\
PUMA \cite{puma} & TIL, CIL & Transductive & Condensation with pseudo labels via distribution matching & Node Classification \\
DSLR \cite{DSLR} & CIL & Unknown & \makecell[l]{Diversity-based node sampling \\ Learns structure adaptively} & Node Classification \\
E-CGL \cite{E-CGL} & TIL & Transductive & Combines PageRank and diversity sampling & Node Classification \\
DeLoMe \cite{DeLoMe} & TIL, CIL & Inductive & Condenses graphs via gradient trajectory alignment & Node Classification \\
FTF-ER \cite{FTF-ER} & CIL & Inductive & Node importance via gradient norm and Hodge score & Node Classification \\
PDGNN \cite{PDGNN} & TIL, CIL & Transductive & Prompt-guided pseudo-instance generation & Node Classification \\
CLDNE \cite{CLDNE} & DIL,TIL & Inductive & \makecell[l]{Uses coexisting nodes for distillation guidance} & \makecell[l]{Node Classification \\ Link Prediction} \\
TAP \cite{TAP} & CIL & Inductive & \makecell[l]{Topology-based class augmentation \\ Prototype calibration} & Node Classification \\
GASIM \cite{gasim} & CIL & Inductive & \makecell[l]{Topology-based class augmentation \\ Prototype calibration} & Node Classification \\
TPP \cite{TPP} & CIL & Transductive & Task-specific prompts & Node Classification \\
\hline
\end{tabular}
\end{adjustbox}
\label{tab:cgl_summary_abbr}
\end{table}
\section{Methods}
\label{method}

Continual Graph Learning (CGL) methods are commonly grouped into regularization-based, replay-based, architecture-based, and hybrid approaches, as illustrated in Figure~\ref{timeline}. Early studies (2020–2021) mostly focused on regularization, but more recent work increasingly favors replay and architecture-based strategies. This shift addresses challenges in class-incremental learning, where simple constraints have proven insufficient (see Section~\ref{analysis}). As a result, researchers are moving toward more selective and scalable solutions across these categories. In regularization-based methods, there has been a transition from applying broad global constraints (e.g., GraphSail and TWP \cite{GraphSail, TWP}) to using more targeted preservation techniques that help maintain adaptability. For instance, DyGRAIN \cite{DyGRAIN} employs selective distillation, and SSRM \cite{SSRM} aligns distributions to preserve important knowledge. Replay-based methods have also progressed, evolving from selecting basic central nodes (e.g., ER-GNN \cite{ERGNN}) to leveraging topological information \cite{SSM, SEM} and employing generative models or graph condensation \cite{SGNN-GR, CaT, puma, DeLoMe} to better manage storage and privacy concerns. At the same time, architecture-based methods focus on isolating parameters specific to each task. While earlier approaches like TPP \cite{TPP} relied on expanding prompts, more recent works such as HPN \cite{HPN} and PI-GNN \cite{PI-GNN} highlight dynamic parameter reuse and reliable identification to enhance scalability. Taken together, recent advances in CGL indicate a move toward nuanced strategies that carefully balance stability, adaptability, and memory limitations.

\subsection{Regularization-based Methods}
Regularization-based methods can be further divided into weight regularization and function regularization. Weight regularization assigns importance scores to parameters based on previous tasks, while function regularization often utilizes knowledge distillation \cite{kd}, where a teacher model guides a student model to retain prior knowledge.

\subsubsection{Weight Regularization}
\textbf{TWP} \cite{TWP} imposes two regularization penalties on the network parameters: one related to the loss, similar to MAS \cite{mas}, and another related to topological information. The topological knowledge is encoded as the sum of interactions between neighboring node representations. Specifically, the neighbor interaction is defined as the attention score for GAT, or as the cosine similarity between neighboring representations, normalized across edges, for other GNN backbones. \textbf{LKGE} \cite{LKGE} tackles the stability-plasticity trade-off in continual knowledge graph embedding by introducing a masked autoencoder with TransE-style encoders \cite{transE}. It updates knowledge non-disruptively without rigid parametric constraints by fusing historical and emerging node representations via weighted summation. The regularization weight for each node is based on the ratio of new to historical neighbors, assigning higher importance to nodes with richer past connectivity. This strategy helps preserve prior knowledge while integrating new information. However, the reliance on TransE limits expressiveness, potentially constraining performance in complex relational settings.

Other works, such as \cite{CGNN, DiCGRL, IncreGNN, TrafficStream, MSCGL}, also leverage weight regularization. Details are discussed in Section \ref{hybrid}.

\subsubsection{Function Regularization}

\textbf{GraphSail} \cite{GraphSail} tackles the forgetting problem in incremental recommendation systems by introducing three distillation components: Local Structure Distillation (LSD), Global Structure Distillation (GSD), and Self-Embedding Distillation (SED). LSD preserves user-item interactions by maintaining the dot product between a node’s embedding and the average embeddings of its neighbors across both previous and current time frames. GSD first identifies the cluster center of item and user node respectively. Then the global position of each node is encoded as its probability distribution belonging to each cluster (The probability distribution is computed according to its node representation and all the cluster centers). A Kullback-Leibler (KL) divergence loss between the positions from previous and current time frames is minimized to retain the node’s global position. Finally, an additional distillation loss is applied to each node to control shifts of its representation. \textbf{RieGrace} \cite{RieGrace} argues that graph-structured data inherently reside in non-Euclidean spaces, embedding them in hyperbolic spaces is more appropriate, which is a gap in existing research (ER-GNN \cite{ERGNN}, HPN \cite{HPN}). In addition, it highlights the importance of label-free CGL, since obtaining labels for continuously arriving graphs is often impractical, an issue often overlooked in the literature (e.g., MSCGL \cite{MSCGL}, TrafficStream \cite{TrafficStream}). To tackle the first challenge, RieGrace introduces AdaRGCN, a novel GNN backbone that can operate in spaces with varying curvature. AdaRGCN includes a neural curvature adapter that estimates the curvature of the incoming graph and embeds it in the appropriate hyperbolic space. To address the second challenge, RieGrace proposes Label-free Lorentz Distillation, which consists of an intra-distillation module and an inter-distillation module. The intra-distillation module encourages the student model to learn from its own representations under different augmentations, improving task performance. The inter-distillation module aligns the current model’s representations with those of the previous model on old tasks, helping to mitigate catastrophic forgetting. Since distillation is performed through contrastive learning in hyperbolic space, a Lorentz projection technique is introduced to enable comparison of embeddings across spaces with varying curvature. \textbf{SSRM} \cite{SSRM} addresses inductive graph incremental learning, diverging from the prevalent transductive setting studies by exsiting works (CGNN \cite{CGNN}, ER-GNN \cite{ERGNN}), where new nodes and edges extend an existing graph, causing structural shifts that alter the neighborhoods of previously seen nodes. Such changes can shift the input distribution, degrading GNN performance on old tasks due to reliance on structural context.. Unlike prior work that often assumes static or separate graphs, SSRM makes GNNs robust to this evolving topology. It introduces a graph-structure-aware regularizer that aligns node representations before and after the graph changes. Specifically, SSRM adds a Maximum Mean Discrepancy (MMD) loss term that penalizes divergence between old nodes’ embeddings computed on the previous and updated graphs, keeping their representations stable despite new connections. It also aligns new node embeddings with old ones to maintain consistency. As a plugin, SSRM complements any incremental graph learning framework by mitigating structural drift without preventing parameter updates directly. \textbf{IncDE} \cite{IncDE} addresses continual knowledge graph embedding (CKGE), focusing on preserving relational structure as new triples are introduced. Unlike prior methods that treat all facts uniformly, IncDE imposes a hierarchical learning order based on graph structure, introducing highly connected entities before isolated ones. It further employs incremental distillation, transferring entity embeddings across layers to retain old knowledge as new facts are integrated. A two-stage training process separates knowledge integration from fine-tuning to avoid degrading prior embeddings. As a regularization-based method, IncDE leverages graph-specific cues to reduce forgetting, highlighting the need for CKGE-specific continual learning strategies.

The works \cite{DyGRAIN, PI-GNN} also employ functional regularization. Further details are provided in Section \ref{hybrid}.

\subsection{Replay-based}
Replay-based methods can be divided into two categories: experience replay, which selects raw samples from past data, and generative replay, which synthesizes pseudo-samples.

\subsubsection{Experience Replay}

\textbf{ERGNN} \cite{ERGNN} argues that traditional continual learning (CL) methods often ignore graph topology. ERGNN uses replay-based strategies for graph neural networks in node classification tasks and introduces three node selection strategies: Mean of Feature (MF), Coverage Maximization (CM), and Influence Maximization (IM). MF selects nodes closest to category centers based on feature mean. CM selects nodes with fewer neighbors from other categories within a fixed distance. IM selects nodes with the highest influence, measured by the parameter shift from their removal. While ERGNN proposes three complementary strategies, it treats graph data as independent samples, neglecting topology. \textbf{SAER} \cite{SAER} preserves users' long-term preferences in continual recommendation tasks. SAER introduces two interaction selection strategies: 1. Uniform Sampling: Randomly selects interactions for broad coverage. 2. Inverse-Degree Sampling: Prioritizes interactions involving users with fewer interactions to address the long-tail problem. The selection probability is defined as: $p{(u, i)} = ({1/d^{\tau}_u})/ (\sum_{(\hat{u}, \hat{i}) \in \mathcal{H}} 1/d^{\tau}_{\hat{u}})$ where $\tau$ is a temperature parameter, $d_u$ represents the degree of user $u$, and $\mathcal{H}$ is the set of interactions. SAER's inverse-degree sampling is tailored for recommendation tasks, limiting its generalizability to other continual graph learning (CGL) scenarios. \textbf{SSM} \cite{SSM}: SSM inspects the memory sparsity problem of previous replay-based methods (e.g., ERGNN) and the challenges of class-incremental CGL tasks. To enhance local structural information, SSM increases subgraph density by including neighbors of selected nodes. The method operates as follows: 1. Select a set of center nodes. 2. For each center node, add a fixed number of 1-hop neighbors to the memory buffer. 3. Expand the neighbor set iteratively. The iteration count $l$ determines the inclusion of $l$-hop neighbors for each center node. SSM provides a straightforward yet impactful solution and highlights the importance of leveraging topological properties in memory buffer design. \textbf{SEM} \cite{SEM} extends the work of SSM \cite{SSM}. It introduces an efficient curvature surrogate based on lazy random walks, selecting important neighbors of each center node. \textbf{TACO} \cite{TACO} argues previous works such as SSM \cite{SSM} fail to adequately capture topological information. It preserves topological information and captures correlations between old and new tasks in continual graph learning. TACO uses a graph coarsening strategy to merge redundant nodes, generating a compact subgraph. It calculates cosine similarity between connected node pairs, ranks edges by similarity, and iteratively merges nodes with the highest similarity until memory capacity is reached. Merged node labels are determined by majority vote. To prevent class disappearance, TACO integrates representative node selection using strategies from \cite{chaudhry2019continual}. TACO aligns co-existing nodes in the memory bank with the current graph by combining the memory buffer subgraph with the current graph during training. TACO effectively captures inter-task relationships, but edge removal during coarsening results in some information loss. \textbf{FTF-ER} \cite{FTF-ER} argues that previous works's (e.g., SSM \cite{SSM}) focus on features or topology hinders comprehensive graph exploitation. It aims to efficiently leverage both feature information and global topological information, where the latter is often overlooked by previous methods. For each node, it computes two scores: one based on node features and the other based on global topology. The feature-based importance is quantified by the relative loss reduction that the node achieves compared to other samples. The global topological importance is measured using a novel Hodge Potential Score (HPS) module, which applies Hodge decomposition to derive a global ranking of nodes by their structural significance, capturing topological properties such as flow and connectivity patterns. By combining these two perspectives, FTF-ER selects nodes that are both attribute-significant and topologically critical. \textbf{DSLR} \cite{DSLR} identifies two key issues with previous experience replay methods. First, storing nodes solely based on class representativeness (e.g., proximity to the class mean) can lead to overlap in node representations, which may cause overfitting. Second, merely storing representative nodes might be suboptimal if their neighbors are irrelevant. To address these concerns, it introduces a coverage-based diversity (CD) sampling strategy and integrates graph structure learning (GSL) into the replay process. The CD strategy aims to maximize the coverage of the feature space for each class, ensuring node diversity within the memory bank. GSL modifies the graph structure by rewiring or weighting edges to connect selected nodes with genuinely informative neighbors while filtering out irrelevant ones. In practice, a link prediction module refines the subgraph structure, ensuring that message passing during replay focuses on meaningful relationships. \textbf{E-CGL} \cite{E-CGL} addresses the interdependencies among graphs and the efficiency challenges posed by large-scale graphs. It introduces a graph-specific replay sampling strategy and a parameter-sharing MLP training framework. Specifically, E-CGL samples nodes based on a score that combines node importance (measured by attributed PageRank) and diversity (the novelty of a node’s features relative to its neighbors). This approach ensures that the memory bank retains both influential nodes and a broad coverage of representative patterns. To improve efficiency, E-CGL trains a lightweight MLP whose weight space is shared with a GCN. During training, only the MLP parameters are updated, without any message passing. The trained MLP parameters are then used to initialize the GCN filters, enabling message passing during inference. \textbf{TAP} \cite{TAP} addresses the challenge of Graph Few-Shot Class-Incremental Learning (GFSCIL) in an inductive setting, where each incremental session introduces a new, disjoint graph with few labeled nodes and no access to previous graphs. This setting introduces severe forgetting and label sparsity. TAP aims to enable adaptation to new classes while retaining old knowledge under strict memory constraints. The method includes three key components: Multi-Topology Class Augmentation, which simulates future incremental scenarios by dividing the base graph into diverse subgraphs during training to improve generalization across isolated structures. Iterative Prototype Calibration, which refines new class prototypes over successive updates to ensure separation and stability despite few-shot data, and Prototype Shift Compensation, which adjusts old class prototypes to account for representation drift as the model is fine-tuned. By focusing on local subgraph features and storing only class prototypes, TAP enables memory-efficient, robust learning in open-world, evolving graph settings. 

The works \cite{CGNN, DiCGRL, IncreGNN, TrafficStream, DyGRAIN} also employ experience replay strategies, with details elaborated in Section \ref{hybrid}.

\subsubsection{Generative Replay}
\textbf{SGNN-GR} \cite{SGNN-GR} overcomes the challenges of previous regularization-based methods (TWP \cite{TWP}) and the storage limitations of experience replay (CGNN \cite{CGNN}, ER-GNN \cite{ERGNN}, FGN \cite{FGN}, TrafficStream \cite{TrafficStream}. It introduces a generative model that learns node topological information using random walks with restart (RWR), which captures neighborhood data. A GRU-based generator creates pseudo walk traces containing node identifiers and attributes, which are evaluated by a discriminator following the WGAN framework \cite{WGAN}. As new data arrives, SGNN-GR quantifies node influence by measuring representation shifts, retraining nodes with significant shifts and eliminating redundant ones. It is the first to apply generative replay in continual graph learning (CGL), addressing knowledge consolidation, new pattern detection, and redundancy elimination. However, it incurs higher training costs due to the GRU model and does not use label information. \textbf{CaT} \cite{CaT} argues that previous replay-based methods such as ER-GNN \cite{ERGNN} and SSM \cite{SSM} fail to preserve sufficient information with limited memory buffer. To address this problem, it condenses the original graph into a small but informative subgraph for storage and replay. To preserve the original distribution, CaT compares node representations from the original and condensed graphs using a Maximum Mean Discrepancy (MMD) loss with a linear kernel, ensuring that intra-class distributions remain aligned. To handle imbalance and reduce training cost, CaT first condenses the new graph into a synthetic subgraph, which is then trained jointly with the memory buffer. Unlike conventional generative replay models, GC produces a subgraph directly, avoiding heavy encoder-decoder overhead. However, optimizing the subgraph can still be computationally demanding, especially for high-dimensional features. \textbf{PUMA} \cite{puma} is an extension of CaT. Frist, it leverages pseudo-labeling to improve predicting performance. Second, it disentangles feature transformation and feature aggregation in node representation learning. Specifically, feature aggregation is performed via a one-step random walk using a normalized Laplacian matrix $L$, calculated as $F = LX$. Feature transformation is handled by a multilayer perceptron (MLP). \textbf{PDGNN} \cite{PDGNN} argues that existing memory replay methods (e.g., ER-GNN \cite{ERGNN}, FGN \cite{FGN}, SSM \cite{SSM}, SEM \cite{SEM}) mitigate catastrophic forgetting in graph learning but face a memory explosion problem. Replaying $n$ nodes in graph neural networks (GNNs) requires storing their $L$-hop neighborhoods, leading to $O(n d^L)$ memory usage, which is unmanageable for large graphs. Existing methods like ER-GNN store only node features and ignore topology, while SSM sparsifies subgraphs by removing many edges and nodes. However, both approaches sacrifice important structural information. PDGNN solves this by decoupling the GNN parameters from specific nodes and using a compact Topology-aware Embedding (TE) memory. Instead of storing full subgraphs, PDGNN embeds each subgraph into a fixed-size TE vector, which captures all the relevant information for model retraining. This reduces memory complexity to $O(n)$, independent of degree $d$ or hop length $L$. During replay, PDGNN feeds the stored TEs into the model's classifier, bypassing the need for neighborhood reconstruction. Additionally, a pseudo-training effect occurs as the GNN's neighbors influence each other's representations. This effect is quantified by a `coverage ratio' which measures the subgraph coverage by a TE. PDGNN leverages a coverage maximization strategy, prioritizing nodes whose $L$-hop subgraphs cover more nodes, thus improving knowledge retention and enhancing memory efficiency. \textbf{DeLoMe} \cite{DeLoMe} critiques experience replay methods (e.g., ER-GNN \cite{ERGNN}, SSM \cite{SSM}) for failing to preserve holistic graph information. It also highlights challenges related to data privacy and class imbalance. To retain comprehensive graph information, DeLoMe employs a graph condensation algorithm, similar to CaT \cite{CaT} and PUMA \cite{puma}. However, instead of distribution matching, it adopts one-step gradient matching \cite{onestep}, which aligns the gradient trajectories between the original graph and the synthetic graph. The condensed subgraph is then trained jointly with new data. To address class imbalance, DeLoMe incorporates a debiased memory replay approach. This approach modifies the prediction logits for both memory and current graph data, considering class label frequencies during replay. \textbf{GRE-FL} \cite{GRE-FL} tackles catastrophic forgetting in federated continual graph learning, where decentralized clients collaboratively train a global GNN without sharing raw data. A key challenge is global forgetting, new client updates may overwrite knowledge from earlier clients. GRE-FL addresses this via a server-side generative replay module that synthesizes a summary graph to retain salient historical information across clients. Additionally, each client employs a gating graph attention network to enhance feature extraction locally. Extending replay-based methods to the federated setting, GRE-FL mitigates both inter-client and inter-task forgetting. While training the generative model introduces overhead, the approach effectively preserves knowledge under distributed and privacy-constrained conditions.

\subsection{Architecture-based}

Some architecture-based methods decompose a fixed large network into several components tailored for different tasks, while others dynamically expand or compress the network to accommodate pattern shifts, resulting in adaptive architectures.

\subsubsection{Fixed Architecture}
Several studies \cite{DiCGRL, FGN, MSCGL} employ architecture-based strategies with fixed architectures. Detailed discussions can be found in Section \ref{hybrid}.

\subsubsection{Dynamic Architecture}

\textbf{HPN} \cite{HPN} aims to learn new classes and their connections without degrading performance on previously learned classes, while keeping memory usage bounded as tasks accumulate. It introduces a hierarchical prototype-based representation that captures knowledge at three levels of abstraction, enabling selective updates. Specifically, it employs a set of Atomic Feature Extractors (AFEs) to decompose each node’s information into its attribute features and local topological context. Based on this, HPN constructs three levels of prototypes: atomic-level prototypes for basic attribute patterns, node-level prototypes for node-specific patterns, and class-level prototypes for broader class patterns. When a new node arrives, only the relevant prototypes are activated and refined, while unrelated prototypes remain unchanged. The refined prototypes are then fused with the new node’s information to support inference. Unlike experience replay methods, the prototype memory grows sub-linearly with the number of tasks, since new classes can reuse existing prototypes. \textbf{PI-GNN} \cite{PI-GNN} addresses the limitations of existing CGL approaches that rely on a single GNN to learn all tasks, which often suffer from the stability–plasticity dilemma. It adopts a parameter isolation and expansion strategy. For each task, PI-GNN identifies stable parts (unaffected by the new graph) and unstable parts (perturbed by the new graph) within the model’s learned knowledge. It then determines which parameters correspond to these stable and unstable components. Parameters related to the stable parts are frozen during training on the new data, ensuring that previous knowledge remains intact. Parameters associated with the unstable parts are updated to accommodate changes in the perturbed regions of the previous graph. Simultaneously, the model expands with new parameters to capture patterns specific to the new graph. \textbf{DEGC} \cite{DEGC} addresses the recommendation problem. Existing CGL methods (e.g., GraphSail \cite{GraphSail}) tend to preserve outdated short-term preferences, which can lead to an overly stable model that fails to adapt new knowledge. DEGC tackles this by dynamically pruning and expanding GNN filters to separate long-term and short-term preferences. Specifically, it prunes parameters that capture obsolete short-term preferences and refines parameters that preserve long-term user interests. In addition, it expands the network with new parameters to capture emerging preferences. To further adapt to user preference drift over time, DEGC incorporates a temporal attention module that models temporal dynamics and adaptively initializes user embeddings. \textbf{GASIM} \cite{gasim} addresses continual graph neural architecture search, where sequential tasks require evolving GNN architectures. Standard NAS methods struggle in this setting due to architecture conflicts, optimal architectures for new tasks often degrade performance on previous ones. GASIM resolves this by disentangling architecture search into modular components, allowing flexible adaptation without forgetting. It introduces a modular supernet, where each module specializes in different task types, and a routing mechanism assigns tasks to modules based on latent graph factors. Within each module, GASIM performs NAS with an invariance loss to preserve performance on previously routed tasks. This design enables continual, memory-efficient architecture adaptation while maintaining task-specific performance. \textbf{TPP} \cite{TPP} introduces a novel prompt-based framework for class-incremental learning on graphs, addressing the challenge of catastrophic forgetting without requiring rehearsal or task identifiers. It maintains a pool of learnable prompts and dynamically assembles task-relevant ones at test time using a context encoder, allowing flexible generalization to new tasks. A class-wise contrastive objective ensures prompt diversity and discriminability, while a prompt distillation strategy transfers knowledge across tasks to prevent forgetting. TPP's lightweight design and prompt-centric approach make it effective for scalable, rehearsal-free continual learning, with strong empirical performance across multiple benchmarks.

\subsection{Hybrid}

\label{hybrid}

\textbf{CGNN} \cite{CGNN} addresses scalability and forgetting in streaming graph data using a combination of replay and regularization. It maintains two memory buffers: one captures nodes with significant representation shifts, while the other retains representative nodes based on class boundaries. \textbf{FGN} \cite{FGN} transforms the CGL problem into a traditional CL problem, enabling the use of conventional CL strategies. It reformulates node classification as graph classification, enabling incremental mini-batch training. Specifically, for each node, a feature graph is constructed where the components of its feature vector become nodes, and edges are formed based on cross-correlations with neighbor features. This converts each node into an independent graph sample, allowing the use of standard continual learning methods such as rehearsal and regularization. New node arrivals are thus treated as new training samples, avoiding the need to update a global graph. A graph convolutional network is applied to these feature graphs, combined with methods like EWC and replay. Despite the independence of samples, the use of feature correlation preserves essential local structure. \textbf{DiCGRL} \cite{DiCGRL} addresses continual learning in knowledge graph completion and node classification by combining parameter isolation and experience replay. It decomposes node embeddings into semantic components, updating only those deemed relevant via attention scores. A neighbor activation replay module selects 1-hop and 2-hop neighbors from past graphs to preserve historical knowledge. Parameter isolation mitigates forgetting, while replayed nodes help integrate new patterns and remove outdated ones. However, further strategies are needed for stronger knowledge consolidation. \textbf{IncreGNN} \cite{IncreGNN} addresses a limitation in previous CGL methods (e.g., CGNN \cite{CGNN}), which risk losing local structural information during probability computations. It specifically targets the continual recommendation problem. IncreGNN combines weight regularization and experience replay. The weight regularization module applies an elastic penalty on model parameters, following \cite{mas}. The experience replay stores two types of nodes: 1. Affected Nodes: Primarily neighbors of new data that undergo significant perturbations due to message propagation along edges. 2. Unaffected Nodes: Nodes not impacted by new data are clustered using the K-Means algorithm, with node degree serving as an importance metric (higher degree indicates greater importance). Similar to CGNN \cite{CGNN}, the replay module first removes outdated information and integrates emerging patterns, then consolidates knowledge to maintain performance across tasks. \textbf{TrafficStream} \cite{TrafficStream} addresses continual graph learning (CGL) for traffic flow prediction, a domain where both node attributes and graph topology evolve over time, a challenge that previous CGL methods (e.g., CGNN \cite{CGNN}, FGN \cite{FGN}) have not adequately tackled. In traffic networks, nodes represent road sensors, and node attributes correspond to the traffic flow recorded over specific periods. It combines two key components: Traffic Pattern Fusion, which uses Jensen–Shannon divergence to detect distribution shifts and retrains on localized subgraphs for new or changing nodes, and Continual Learning Strategies, which employ experience replay to retain past knowledge and parameter regularization to prevent forgetting. This allows the model to adapt to dynamic traffic networks while maintaining accuracy over time. \textbf{Trans-CGL} \cite{trans-CGL} tackles topological-feature-induced catastrophic forgetting (TCF) in continual graph learning, where models forget not only node features but also shared structural patterns across tasks. Unlike prior methods (e.g., CGNN \cite{CGNN}, TWP \cite{TWP}) focused on node-level forgetting, Trans-CGL leverages the high capacity of Transformers to preserve both node and topological knowledge. It introduces two key components: Gated Attention via Parameter Isolation, which uses task-specific binary masks to activate disjoint subsets of attention parameters, isolating old task representations, and Asymmetric Masked Attention Regularization, which constrains shared attention weights to prevent forgetting, penalizing harmful drift more than restricting plasticity. This design allows the model to retain useful patterns from past tasks while adapting to new ones, achieving strong performance in sequential graph learning scenarios. \textbf{DyGRAIN} \cite{DyGRAIN} addresses continual learning in dynamic graphs, where evolving structures introduce new challenges beyond standard forgetting. In particular, new nodes or edges can shift the receptive fields of existing nodes, altering their embeddings and leading to incorrect predictions, even without parameter updates. This time-varying receptive field problem, combined with catastrophic forgetting, which is a research gap of previous literature (e.g., GraphSail \cite{GraphSail}, TWP \cite{TWP}), motivates DyGRAIN’s design. The framework includes two components: Influence Propagation, which identifies old nodes affected by new graph changes through message passing and updates their embeddings accordingly, and Knowledge Distillation with Truth, which selectively reinforces true labels for old nodes whose predictions degrade after updates. By focusing on structurally impacted and vulnerable nodes, DyGRAIN preserves prior knowledge while adapting to new graph data. \textbf{MSCGL} \cite{MSCGL} aims to integrate multimodal information into continual graph learning (CGL) and to capture evolving structural patterns. It introduces an Adaptive Multimodal GNN (AdaMGNN) whose architecture is continually optimized using Neural Architecture Search (NAS) in conjunction with Group Sparse Regularization (GSR) \cite{scardapane2017group, pasunuru2019continual} to preserve important parameters. Specifically, as a new task arrives, MSCGL performs joint NAS and GSR optimization. NAS searches for the optimal GNN architecture for the new task (e.g., selecting appropriate aggregators and activation functions), while GSR imposes group sparsity on parameters to isolate and preserve important weights from previous tasks. In practice, the model can expand by adding new GNN cells or operations as needed for the new task, while key weight groups from prior tasks remain sparse (i.e., unchanged) to mitigate forgetting. \textbf{UGCL} \cite{UGCL} tackles the challenge of learning across heterogeneous graph tasks, such as alternating node and graph classification, within a single continual learning framework. Unlike prior methods limited to one task type, UGCL uses a unified GNN architecture and a rehearsal-based strategy to preserve both local and global structural knowledge. The model reuses shared parameters across tasks, with a flexible readout for node- or graph-level outputs. A replay buffer stores node-centric subgraphs and full graphs, while a consistency loss ensures past representations remain stable. By maintaining structural consistency across task types, UGCL enables effective continual learning in mixed-task settings. \textbf{Mecoin} \cite{Mecoin} addresses graph class-incremental learning under extreme few-shot settings by integrating external memory with knowledge preservation. It introduces Structured Memory Units to store compact class prototypes and Memory Construction Modules to update them as new classes arrive. A key innovation is the Memory Representation Adaptation Module, which models each prototype via probability distributions over feature patterns, avoiding full GNN fine-tuning. During new class learning, relevant prototypes are retrieved, and Graph Knowledge Distillation transfers their compressed knowledge back into the model, mitigating forgetting. Mecoin combines elements of replay-based and regularization-based approaches, offering efficient retention of prior knowledge with minimal samples, a critical need in graph domains with sparse labels. While managing multiple memory modules adds complexity, Mecoin achieves low forgetting and strong accuracy, highlighting the value of hybrid strategies for few-shot incremental graph learning. \textbf{CLDNE} \cite{CLDNE} addresses continual dynamic network embedding by preserving global structural patterns over time, an aspect often overlooked in prior work (e.g., SGNN-GR \cite{SGNN-GR}). It combines two components: an experience replay module, which performs $k$-hop random walks from nodes in historical and current graphs and prunes redundant edges to form a compact subgraph, and a knowledge distillation module to retain prior knowledge. The merged subgraph and new data are input into a graph autoencoder, which learns embeddings by minimizing reconstruction loss, preserving both global structure and local proximity. Instead of replaying memory samples directly, CLDNE uses them to guide unsupervised learning, enabling efficient and memory-aware representation updates.

\subsection{Relating CGL Methods to Specific Use-cases}

The effectiveness of continual graph learning (CGL) methods depends strongly on the characteristics of the target application. In node classification tasks with relatively stable label semantics, regularization-based approaches such as EWC and TWP perform well under the Task-IL setting, where task identifiers and disjoint label spaces are available \cite{ewc, TWP}. However, under the more challenging Class-IL setting, these methods often suffer performance degradation, and alternative strategies such as replay-based methods (e.g., CaT, PUMA) or prompt-based models (e.g., TPP) generally demonstrate better adaptability \cite{TPP}. In dynamic scenarios such as recommendation and traffic prediction, where data distributions shift over time, methods that incorporate adaptive architectures or distribution-aware memory construction are more appropriate. For instance, DEGC selectively forgets short-term user preferences while retaining long-term patterns, and TrafficStream emphasizes the replay of nodes exhibiting significant distributional shifts \cite{DEGC, TrafficStream}. In contrast, methods such as GraphSAIL may become overly conservative by preserving outdated patterns, thus limiting adaptability to recent changes \cite{GraphSail}.

In domains such as knowledge graph embedding and recommendation, where new entities and relations continuously emerge (commonly under Domain-IL), models like DiCGRL employ parameter isolation and selective replay to support efficient adaptation while preserving past knowledge \cite{DiCGRL}. The assumptions regarding graph access also play a critical role. Some methods, such as TPP, are designed for transductive settings where the entire graph is available during both training and inference \cite{TPP}. However, in inductive scenarios where future data or structures may not be seen in advance, it is preferable to adopt models that can operate without full-graph access. The graph structure also affects how tasks are related. When edges exist between tasks,  it is critical to utilize these connections to facilitate knowledge transfer and mitigate forgetting \cite{CGNN, DiCGRL, PI-GNN, IncreGNN}. If such cross-task edges are unavailable, it becomes important to retain structural patterns within each task, so that essential topological information is not lost over time.
% \begin{figure*}[ht]
% \centering
%     \hspace{5mm} 
% 	\subfloat[Task-incremental]{\includegraphics[width = 0.25\textwidth]{CGL_taskIL.png}}
% 	\label{fig3:a}
% 	\hfill
% 	\subfloat[Domain-incremental]{\includegraphics[width = 0.25\textwidth]{CGL_domainIL.png}}
% 	\label{fig3:b}
% 	\hfill
% 	\subfloat[Class-incremental]{\includegraphics[width = 0.35\textwidth]{CGL_classIL.png}} 
% 	\label{fig3:c}
% 	 \hspace{5mm} 
	
% \caption{Three scenarios of CGL.}
% \label{fig:CGL_cat}
% \end{figure*}

\begin{table}[htp]
\centering
\caption{Summary of Datasets Used in CGL Literature. NC, NR, KGC, IR, and GC represent Node Classification, Node Regression, Knowledge Graph Completion, Incremental Recommendation, and Graph Classification, respectively.}
\label{dataset}
\scriptsize
\begin{adjustbox}{max height=\textheight, max width = \textwidth}
\begin{tabular}{c|c|c|c|c}
\hline
Granularity                  & Task                & Dataset      & Source                                                           & Literature                                                                                                                       \\ \hline
\multirow{17}{*}{Node-level} & \multirow{16}{*}{NC} & Cora \cite{cora}          & linqs.soe.ucsc.edu/data                                          & \begin{tabular}[c]{@{}c@{}} \cite{CGNN}, \cite{ERGNN}, \cite{FGN},\\ \cite{DiCGRL}, \cite{HPN}, \cite{RieGrace},\\  \cite{trans-CGL}, \cite{PI-GNN}, \cite{DSLR} \end{tabular}                                   \\ \cline{3-5} 
                             &                      & Citeseer \cite{cora}      & linqs.soe.ucsc.edu/data                                          & \begin{tabular}[c]{@{}c@{}} \cite{ERGNN}, \cite{FGN},\cite{DiCGRL},\\ \cite{HPN}, \cite{RieGrace}, \cite{trans-CGL},\\ \cite{PI-GNN}, \cite{DSLR}\end{tabular}                                        \\ \cline{3-5} 
                             &                      & Amazon \cite{amazon}        & nijianmo.github.io/amazon/index.html                             & \begin{tabular}[c]{@{}c@{}} \cite{TWP}, \cite{MSCGL}, \cite{puma},\\ \cite{trans-CGL}, \cite{PI-GNN}, \cite{FTF-ER},\\ \cite{DSLR}\end{tabular}                                                \\ \cline{3-5} 
                             &                      & Elliptic \cite{elliptic}      & www.kaggle.com/datasets/ellipticco/elliptic-data-set             & \cite{CGNN}, \cite{PI-GNN}                                                                                                                              \\ \cline{3-5} 
                             &                      & PPI \cite{ppi}          & snap.stanford.edu/biodata/datasets/10000/10000-PP-Pathways.html  & \cite{TWP}                                                                                                                                      \\ \cline{3-5} 
                             &                      & Corafull \cite{corafull}      & github.com/shchur/gnn-benchmark\#datasets                        & \begin{tabular}[c]{@{}c@{}} \cite{TWP}, \cite{SSM}, \cite{SEM},\\ \cite{puma}, \cite{CaT}, \cite{E-CGL},\\ \cite{DeLoMe}, \cite{FTF-ER}, \cite{PDGNN}\end{tabular}                                        \\ \cline{3-5} 
                             &                      & Reddit \cite{reddit}        & www.reddit.com/r/datasets/                                       & \begin{tabular}[c]{@{}c@{}} \cite{TWP}, \cite{ERGNN}, \cite{RieGrace},\\ \cite{SSM}, \cite{DyGRAIN}, \cite{SEM},\\ \cite{puma}, \cite{CaT}, \cite{E-CGL},\\ \cite{DeLoMe}, \cite{FTF-ER}, \cite{PDGNN},\\  \cite{DSLR} \end{tabular}      \\ \cline{3-5} 
                             &                      & Actor \cite{actor}        & github.com/graphdml-uiuc-jlu/geom-gcn/tree/master/new\_data/film & \cite{HPN}, \cite{RieGrace}                                                                                                                             \\ \cline{3-5} 
                             &                      & Pubmed \cite{pubmed}        & pubmed.ncbi.nlm.nih.gov/download/                                &  \cite{FGN}, \cite{DiCGRL}, \cite{DyGRAIN}                                                                                                                       \\ \cline{3-5} 
                             &                      & DBLP \cite{dblp}          & www.aminer.cn/data/?nav=openData                                 & \cite{CGNN}, \cite{PI-GNN}, \cite{TACO}                                                                                                                         \\ \cline{3-5} 
                             &                      & WikiCS \cite{wikics}        & github.com/pmernyei/wiki-cs-dataset                              & \cite{trans-CGL}                                                                                                                                \\ \cline{3-5} 
                             &                      & OGB-Paper100M  & ogb.stanford.edu/docs/nodeprop/\#ogbn-papers100M                 &  \cite{PI-GNN}                                                                                                                                   \\ \cline{3-5} 
                             &                      & OGB-Arxiv     & ogb.stanford.edu/docs/nodeprop/\#ogbn-arxiv                      & \begin{tabular}[c]{@{}c@{}} \cite{FGN}, \cite{HPN}, \cite{RieGrace},\\ \cite{SSM}, \cite{DyGRAIN}, \cite{SSM},\\  \cite{puma}, \cite{CaT}, \cite{PI-GNN},\\  \cite{E-CGL}, \cite{DeLoMe}, \cite{FTF-ER},\\ \cite{PDGNN}, \cite{DSLR}\end{tabular} \\ \cline{3-5} 
                             &                      & OGB-Products  & ogb.stanford.edu/docs/nodeprop/\#ogbn-products                   & \begin{tabular}[c]{@{}c@{}} \cite{HPN}, \cite{SSM}, \cite{DyGRAIN},\\  \cite{SEM}, \cite{puma}, \cite{CaT},\\  \cite{E-CGL}, \cite{DeLoMe}, \cite{PDGNN}\end{tabular}                                       \\ \cline{3-5} 
                             &                      & Kindle \cite{kindle}       & -                                                                & \cite{TACO}                                                                                                                                    \\ \cline{3-5} 
                             &                      & ACM          & www.openicpsr.org/openicpsr/project/100843/version/V2/view       &  \cite{TACO}                                                                                                                                    \\ \cline{2-5} 
                             & NR                   & PEMS3 \cite{PEMS}  & dot.ca.gov/programs/traffic-operations/mpr/pems-source           & \cite{TrafficStream}                                                                                                                            \\ \hline
\multirow{9}{*}{Edge-level}  & \multirow{2}{*}{KGC} & FB15k-237 \cite{fb15k237}    & huggingface.co/datasets/KGraph/FB15k-237                        & \cite{DiCGRL}                                                                                                                                   \\ \cline{3-5} 
                             &                      & WN18RR \cite{wn18rr}       & huggingface.co/datasets/VLyb/WN18RR                              &  \cite{DiCGRL}                                                                                                                                   \\ \cline{2-5} 
                             & \multirow{7}{*}{IR}  & Gowalla      & snap.stanford.edu/data/loc-gowalla.html                          & \cite{GraphSail}, \cite{SAER}                                                                                                                           \\ \cline{3-5} 
                             &                      & LastFM        & grouplens.org/datasets/hetrec-2011/                              &  \cite{GraphSail}                                                                                                                                \\ \cline{3-5} 
                             &                      & Taobao2014    & tianchi.aliyun.com/dataset/53                                    &  \cite{SAER}, \cite{DEGC}                                                                                                                               \\ \cline{3-5} 
                             &                      & Taobao2015    & tianchi.aliyun.com/dataset/53                                    &  \cite{SAER}, \cite{DEGC}                                                                                                                                \\ \cline{3-5} 
                             &                      & Netflix      & github.com/amirtds/kaggle-netflix-tv-shows-and-movies            &  \cite{DEGC}                                                                                                                                     \\ \cline{3-5} 
                             &                      & Foursquare \cite{foursquare1, foursquare2}    & sites.google.com/site/yangdingqi/home/foursquare-dataset                                                                &  \cite{DEGC}                                                                                                                                     \\ \cline{3-5} 
                             &                      & Yelp \cite{graphsaint}       & www.yelp.com/dataset                                             &  \cite{SAER}                                                                                                                                     \\ \hline
Graph-level                  & GC                   & Tox21 \cite{tox21}         & tripod.nih.gov/tox21/challenge/                                             & \cite{TWP}                                                                                                                                      \\ \hline
\end{tabular}
\end{adjustbox}
\end{table}

\section{Datasets, Code, and Benchmarks}

\subsection{Datasets and Open-source Code}

Table \ref{dataset} presents an overview of each dataset, detailing problem granularity, specific task, dataset name, source, and associated literature. Table \ref{code} summarizes the open-source implementations available for current CGL studies. To provide concrete real-world context, five representative node classification datasets used in the experiments are described: CoraFull, Citeseer, Reddit, OGB-Arxiv, and OGB-Products. CoraFull, Citeseer, and OGB-Arxiv are citation networks in which nodes represent academic papers and edges denote citation relationships. The primary objective is to predict the research field of each paper. In the CGL setting, these datasets are typically partitioned by label; for example, the 70 classes in CoraFull are often divided into 35 sequential two-way classification tasks. The Reddit dataset serves as a benchmark for inductive node classification and consists of posts from September 2014. In this dataset, nodes represent posts, and an edge exists between two posts if the same user commented on both. Node features are derived from GloVe word embeddings of the titles and comments. To simulate an evolving environment, the dataset is split based on time, with training on the first 20 days and testing on the remainder. The OGB-Products dataset models an Amazon product co-purchasing network as an unweighted, undirected graph. Nodes correspond to products, and edges connect items that are frequently bought together. Node features are generated by applying Principal Component Analysis (PCA) to bag-of-words representations of product descriptions, resulting in 100-dimensional vectors. For additional details on the remaining datasets, including specific data statistics and open-source links, Table \ref{dataset} provides further information.

\subsection{Benchmarks}
CGLB \cite{CGLB} and Begin \cite{Begin} are two key benchmarks for continual graph learning (CGL). CGLB provides a framework for node-level and graph-level CGL under task-incremental and class-incremental settings, supporting both transductive and inductive settings, with or without inter-task edges. It includes several baselines, such as EWC \cite{ewc}, MAS \cite{mas}, GEM \cite{GEM}, LWF \cite{LWF}, and TWP \cite{TWP}, and offers a pipeline for designing novel methods, including modules for data processing, baseline evaluation, and result visualization. The code is available at github.com/QueuQ/CGLB, implemented using PyTorch and the DGL library. Begin, on the other hand, supports node-level, link-level, and graph-level CGL problems under task-incremental, domain-incremental, and class-incremental settings, and introduces a time-incremental setting, which is useful for incremental recommendation tasks. Begin offers a broader set of baselines, including EWC \cite{ewc}, MAS \cite{mas}, HAT \cite{HAT}, PackNet \cite{mallya2018packnet}, PiggyBack \cite{Piggyback}, GEM \cite{GEM}, LWF \cite{LWF}, TWP \cite{TWP}, CGNN \cite{CGNN}, ERGNN \cite{ERGNN}, and CaT \cite{CaT}. The code is available at github.com/ShinhwanKang/BeGin, implemented using PyTorch, DGL, and PyG libraries.

% Please add the following required packages to your document preamble:
% \usepackage{multirow}
% Please add the following required packages to your document preamble:
% \usepackage{multirow}

\begin{table}[htp]
\centering
\caption{Summary of Open-Source Code Links. Entries marked with * indicate official implementations. $-$ denotes cases where the authors have provided a link but have not yet uploaded the code. Torch, tensorflow, dgl, and pyg refer to PyTorch, TensorFlow, the Deep Graph Library (DGL) \cite{dgl}, and PyTorch Geometric (PyG) \cite{pyg}, respectively. PyTorch and TensorFlow are deep learning frameworks while DGL and PyG are specifically designed for deep graph learning.}
\label{code}
\scriptsize
\begin{tabular}{c|c|c}
\hline
Methods       & Link                                                                                                                                            & Main Tools                                                                    \\ \hline
GraphSail \cite{GraphSail}     & github.com/mmvv11/GraphSAIL*                                                                                                            & torch dgl                                                                     \\ \hline
TWP \cite{TWP}           & \begin{tabular}[c]{@{}c@{}}github.com/hhliu79/TWP*\\ github.com/QueuQ/CGLB\\ github.com/ShinhwanKang/BeGin\end{tabular} & \begin{tabular}[c]{@{}c@{}}torch dgl\\ torch dgl\\ torch dgl pyg\end{tabular} \\ \hline
CGNN \cite{CGNN}          & github.com/ShinhwanKang/BeGin                                                                                                           & torch dgl pyg                                                                 \\ \hline
ERGNN \cite{ERGNN}         & github.com/ShinhwanKang/BeGin                                                                                                           & torch dgl pyg                                                                 \\ \hline
FGN \cite{FGN}           & github.com/sair-lab/LGL*                                                                                                                & torch dgl                                                                     \\ \hline
DiCGRL \cite{DiCGRL}        & github.com/KXY-PUBLIC/DiCGRL                                                                                                         & torch                                                                         \\ \hline
TrafficStream \cite{TrafficStream} & github.com/AprLie/TrafficStream*                                                                                                        & torch pyg                                                                     \\ \hline
SGNN-GR \cite{SGNN-GR}       & github.com/Junshan-Wang/SGNN-GR$^-$                                                                                                      & -                                                                             \\ \hline
LKGE \cite{LKGE}          & github.com/nju-websoft/LKGE*                                                                                                            & torch pyg                                                                     \\ \hline
SSM \cite{SSM}           & github.com/QueuQ/SSM*                                                                                                                   & torch dgl                                                                     \\ \hline
PUMA \cite{puma}          & github.com/superallen13/PUMA*                                                                                                           & torch pyg                                                                     \\ \hline
CaT \cite{CaT}          & github.com/superallen13/CaT-CGL*                                                                                                        & torch pyg                                                                     \\ \hline
PI-GNN \cite{PI-GNN}       & github.com/Jerry2398/PI-GNN*                                                                                                            & torch pyg                                                                     \\ \hline
DEGC \cite{DEGC}       & github.com/BokwaiHo/DEGC*                                                                                                            & tensorflow                                                                     \\ \hline
SSRM \cite{SSRM}          & github.com/jwsu825/NGIL\_Evolve*                                                                                                        & torch dgl                                                                     \\ \hline
E-CGL \cite{E-CGL}         & github.com/aubreygjh/E-CGL*                                                                                                             & torch dgl                                                                     \\ \hline
FTF-ER \cite{FTF-ER}        & github.com/CyanML/FTF-ER*                                                                                                               & torch dgl                                                                     \\ \hline
DSLR \cite{DSLR}         & github.com/seungyoon-Choi/DSLR\_official*                       & torch pyg  \\ \hline
PDGNN \cite{PDGNN}   &  github.com/imZHANGxikun/PDGNNs*              &  torch dgl      \\ \hline
DeLoMe \cite{DeLoMe}       & github.com/QueuQ/CGLB/tree/master*      & torch dgl      \\ \hline
TACO \cite{TACO}       & github.com/hanxiaoxue114/TACO*                    & torch dgl      \\ \hline
TPP \cite{TPP}       & github.com/mala-lab/TPP*                    & torch dgl      \\ \hline
\end{tabular}
\end{table}

\begin{table}[htp]
\centering
\scriptsize
\caption{Comparison result under Task-IL setting using GCN. Reg, Rep, and Arc denotes Regularization-based, Replay-based, and Architecture-base methods respectively. }
\label{result_1}
\resizebox{\textwidth}{!}{
\begin{tabular}{ll|cc|cc|cc|cc|cc}
\hline
\multicolumn{2}{c|}{Method} & \multicolumn{2}{c|}{CoraFull} & \multicolumn{2}{c|}{Citeseer} & \multicolumn{2}{c|}{Arxiv} & \multicolumn{2}{c|}{Reddit} & \multicolumn{2}{c}{Products} \\
\makecell{Family} & Name & AP & AF & AP & AF & AP & AF & AP & AF & AP & AF \\
\hline
\multirow{3}{*}{Reg}
& EWC \cite{ewc} & 51.8±2.0 & -42.8±2.4 & 68.8±3.1 & -22.6±3.7 & 73.5±0.9 & -12.2±1.0 & 61.3±1.6 & -40.2±1.7 & 74.5±1.9 & -20.1±2.1 \\
& TWP \cite{TWP} & 77.3±2.6 & -16.0±2.4 & \underline{78.1}±2.7 & -8.0±4.2 & 82.5±0.9 & -11.8±0.9 & -- & -- & -- & -- \\
& SSRM \cite{SSRM} & 91.7±0.6 & 0.4±0.4 & 82.2±0.2 & \textbf{1.7}±0.6 & 88.4±0.1 & -2.6±0.2 & -- & -- & -- & -- \\
\hline
\multirow{7}{*}{Rep}
& ERGNN-MF \cite{ERGNN}& 91.9±0.1 & -2.5±0.2 & 75.8±0.1 & -12.7±0.3 & 80.8±0.2 & -10.8±0.2 & 49.5±2.2 & -52.6±2.3 & 58.5±0.4 & -36.9±0.4 \\
& ERGNN-CM \cite{ERGNN}& \textbf{94.7}±0.1 & \underline{0.4}±0.1 & \underline{82.6}±0.3 & -1.2 ±0.5 & \textbf{88.9}±0.1 & -2.4±0.1 & \underline{99.3}±0.0 & -0.3±0.0 & 91.6±0.1 & \underline{-3.7}±0.1 \\
& SSM-degree \cite{SSM} & 91.1±0.6 & -2.6±0.5 & 80.0±1.3 & -5.8±2.0 & 79.6±0.5 & -4.3±0.5 & 98.8±0.1 & -0.8±0.1 & 87.1±0.4 & -8.5±0.5 \\
& CaT \cite{CaT} & 92.5±0.2 & 0.2±0.2 & 71.8±0.7 & -0.5±0.9 & 66.3±0.5 & -5.0±0.8 & 98.2±0.1 & -0.0±0.2 & 92.0±0.6 & \textbf{0.4}±0.3 \\
& PUMA \cite{puma} & 90.9±0.3 & \textbf{0.5}±0.3 & 64.7±2.1 & -0.2±1.1 & 73.1±0.3 & -2.4±0.6 & 98.3±0.0 & \underline{0.0}±0.1 & -- & -- \\
& PDGNN \cite{PDGNN} & \underline{94.3}±0.3 & 0.0±0.2 & \textbf{83.5}±1.8 & \underline{1.0}±2.3 & \underline{88.8}±0.2 & \underline{-1.2}±0.4 & 99.0±0.1 & -0.5±0.1 & \underline{92.9}±0.5 & -2.6±0.3 \\
\hline
\multirow{1}{*}{Arc}
& TPP \cite{TPP}& 92.5 ±0.3 & 0.0±0.0 & 75.8±0.0 & 0.0±0.0 & 85.5±0.1 & \textbf{0.0}±0.0 & \textbf{99.5}±0.0 & \textbf{0.0} ±0.0 & \textbf{93.6}±0.6 & \underline{0.0}±0.0 \\
\hline
\end{tabular}}
\end{table}

\begin{table}[htp]
\centering
\scriptsize
\caption{Comparison result under Task-IL setting using SGC.}
\label{result_2}
\resizebox{\textwidth}{!}{
\begin{tabular}{ll|cc|cc|cc|cc|cc}
\hline
\multicolumn{2}{c|}{Method} & \multicolumn{2}{c|}{CoraFull} & \multicolumn{2}{c|}{Citeseer} & \multicolumn{2}{c|}{Arxiv} & \multicolumn{2}{c|}{Reddit} & \multicolumn{2}{c}{Products} \\
\makecell{Family} & Name & AP & AF & AP & AF & AP & AF & AP & AF & AP & AF \\
\hline
\multirow{3}{*}{Reg}
& EWC & 53.3±2.7 & -38.7±4.5 & 72.7±0.9 & -16.1±1.2 & 68.8±4.9 & -32.3±5.1 & 77.3±3.1 & -9.2±3.4 & 78.7±1.2 & -13.6±1.1 \\
& TWP & 82.5±0.9 & -11.8±0.9 & 70.8±0.6 & -17.5±0.8 & 88.4±0.6 & 1.4±0.6 & -- & -- & -- & -- \\
& SSRM & 94.6±0.2 & 1.5±1.3 & 81.8±0.0 & -1.2±0.3 & -- & -- & -- & -- & -- & -- \\
\hline
\multirow{7}{*}{Rep}
& ERGNN-MF & 92.4±0.1 & -0.5±1.7 & 76.6±0.0 & -10.8±0.3 & 76.0±0.1 & -15.4±0.3 & 46.3±1.3 & -56.0±1.4 & 55.5±0.5 & -42.1±0.5 \\
& ERGNN-CM & 94.7±0.2 & \underline{1.8}±1.3 & 81.9±0.1 & -1.1±0.3 & 88.3±0.1 & -2.7±0.2 & 99.3±0.0 & -0.3±0.0 & 92.0±0.6 & -3.5±0.2 \\
& SSM-degree & 92.5±0.4 & 0.4±1.7 & 79.1±0.5 & -6.4±0.9 & 86.6±0.3 & -4.1±0.5 & 98.9±0.1 & -0.7±0.1 & 88.4±0.7 & -7.9±0.5 \\
& CaT & \underline{94.8}±0.4 & 0.7±0.3 & \underline{84.2}±0.6 & \textbf{0.5}±1.1 & \underline{89.7}±0.3 & -0.0±0.4 & 99.2±0.1 & 0.0±0.2 & 90.0±0.8 & -1.0±0.3 \\
& PUMA & \textbf{95.7} ±0.1 & 0.2±0.2 & \textbf{84.4}±0.6 & -0.3±0.8 & \textbf{89.8}±0.3 & -0.3±0.2 & 99.3±0.0 & 0.0±0.0 & -- & -- \\
& PDGNN & 94.2±0.3 & \textbf{2.8}±1.5 & 82.9±1.7 & \underline{0.1}±2.7 & 88.8±0.4 & -1.1±0.6 & 88.8±0.4 & -1.1±0.6 & \underline{92.4}±0.2 & -3.5±0.2 \\
\hline
\multirow{1}{*}{Arc}
& TPP & 92.4±0.0 & 0.0±0.0 & 76.1±0.0 & 0.0±0.0 & 84.5±0.4 & 0.0±0.0 & 95.5±0.0 & 0.0±0.0 & \textbf{93.0}±0.6 & \textbf{0.0}±0.0 \\
\hline
\end{tabular}}
\end{table}

\begin{table}[htp]
\centering
\scriptsize
\caption{Comparison result under Class-IL setting using GCN.}
\label{result_3}
\resizebox{\textwidth}{!}{
\begin{tabular}{ll|cc|cc|cc|cc|cc}
\hline
\multicolumn{2}{c|}{Method} & \multicolumn{2}{c|}{CoraFull} & \multicolumn{2}{c|}{Citeseer} & \multicolumn{2}{c|}{Arxiv} & \multicolumn{2}{c|}{Reddit} & \multicolumn{2}{c}{Products} \\
Family & Name & AP & AF & AP & AF & AP & AF & AP & AF & AP & AF \\
\hline
\multirow{3}{*}{Reg}
 & EWC & 15.4±0.9 & -82.1±1.1 & 31.8±0.1 & -78.2±0.2 & 4.9±0.0 & -90.0±0.3 & 5.0±0.0 & -99.6±0.0 & 6.0±1.0 & -92.0±1.0 \\
 & TWP & 2.8±0.3 & -93.6±0.3 & 31.4±0.1 & -78.8±0.2 & -- & -- & -- & -- & -- & -- \\
  & SSRM & 76.5±0.2 & -14.5±0.3 & 64.3±0.6 & -28.5±1.0 & 16.2±0.5 & -70.7±0.8 & 80.7±0.3 & -19.5±0.3 & -- & -- \\
  \hline
\multirow{7}{*}{Rep}
 & ERGNN-MF & 60.2±0.0 & -32.6±0.1 & 38.3±0.2 & -69.5±0.4 & 10.5±0.2 & -81.8±0.3 & 9.8±0.5 & -94.3±0.5 & 19.1±0.4 & -74.2±0.5 \\
 & ERGNN-CM & 74.0±0.4 & -18.3±0.5 & 57.2±1.6 & -39.9±2.5 & 17.1±0.7 & -71.5±0.8 & 85.7±0.3 & -14.2±0.4 & 59.9±0.4 & -30.4±0.4 \\
 & SSM-degree & 67.0±0.9 & -26.3±0.9 & 43.5±3.6 & -61.4±5.4 & 15.4±0.3 & -75.0±0.4 & 82.7±1.3 & -17.3±1.4 & 53.2±0.5 & -37.5±0.1 \\
 & CaT & 76.3±0.2 & -5.9±0.3 & 73.7±0.5 & \underline{-7.5±0.8} & 22.5±0.2 & \underline{-9.2±0.3} & 97.4±0.0 & \underline{-0.1±0.0} & 60.0±0.6 & \underline{-8.5±0.6} \\
 & PUMA & 72.2±0.3 & -6.0±0.2 & 34.4±2.9 & -16.8±6.4 & 22.6±0.2 & -11.3±0.3 & 97.4±0.0 & -0.1±0.0 & -- & -- \\
 & PDGNN & \underline{81.1±0.1} & \underline{-4.6±0.1} & \underline{75.6±0.1} & -7.8±0.1 & \underline{52.7±0.1} & -12.2±0.2 & \underline{97.6±0.1} & -0.5±0.0 & \underline{75.8±0.2} & -37.5±0.1 \\
 \hline
Arc & TPP & \textbf{93.1±0.5} & \textbf{0.0±0.0} & \textbf{78.2±1.7} & \textbf{0.0±0.0} & \textbf{85.6±0.1} & \textbf{0.0±0.0} & \textbf{99.4±0.0} & \textbf{0.0±0.0} & \textbf{93.4±0.4} & \textbf{0.0±0.0} \\
\hline
\end{tabular}}
\end{table}

\begin{table}[htp]
\centering
\scriptsize
\caption{Comparison result under Class-IL setting using SGC.}
\label{result_4}
\resizebox{\textwidth}{!}{
\begin{tabular}{ll|cc|cc|cc|cc|cc}
\hline
\multicolumn{2}{c|}{Method} & \multicolumn{2}{c|}{CoraFull} & \multicolumn{2}{c|}{Citeseer} & \multicolumn{2}{c|}{Arxiv} & \multicolumn{2}{c|}{Reddit} & \multicolumn{2}{c}{Products} \\
Family & Name & AP & AF & AP & AF & AP & AF & AP & AF & AP & AF \\
\hline
\multirow{3}{*}{Reg} 
& EWC & 7.0±0.1 & -89.8±0.1 & 31.5±0.1 & -79.6±0.1 & 4.9±0.0 & -90.4±0.2 & 5.0±0.1 & -99.6±0.1 & 4.4±0.1 & -96.5±0.1 \\
& TWP & 3.1±0.1 & -93.6±0.1 & 31.6±0.1 & -77.0±0.1 & 4.9±0.0 & -89.7±0.3 & -- & -- & -- & -- \\
& SSRM & 76.6±0.1 & -15.5±0.1 & 65.1±0.1 & -26.9±0.1 & 36.1±0.2 & -46.3±0.1 & 84.5±0.1 & -15.4±0.1 & -- & -- \\
\hline
\multirow{6}{*}{Rep} 
& ERGNN-MF & 59.7±0.1 & -33.3±0.1 & 36.5±0.1 & -70.8±0.1 & 15.9±0.5 & -75.2±0.4 & 11.3±0.1 & -92.6±0.1 & 21.0±0.1 & -73.6±0.1 \\
& ERGNN-CM & 77.1±0.1 & -14.5±0.1 & 59.1±0.1 & -36.3±0.1 & 39.0±0.3 & -47.2±0.3 & 88.1±0.1 & -11.6±0.1 & 62.2±0.1 & -28.1±0.1 \\
& SSM-degree & 67.1±0.1 & -26.1±0.1 & 47.1±0.1 & -55.0±0.1 & 34.2±0.6 & -53.6±1.0 & 86.1±0.1 & -13.7±0.1 & 54.7±0.1 & -36.3±0.1 \\
& CAT & \underline{81.4±0.1} & \underline{-3.7±0.1} & \underline{76.4±0.1} & -9.0±0.1 & \underline{49.5±0.5} & \underline{-12.0±0.5} & 91.8±0.1 & -4.1±0.1 & 56.8±0.1 & -11.6±0.1 \\
& PUMA & 79.8±0.1 & -4.2±0.1 & 75.1±0.1 & -9.8±0.1 & 49.2±0.4 & -12.2±0.9 & 91.5±0.1 & -4.4±0.1 & -- & -- \\
& PDGNN & 78.5±0.1 & -5.4±0.1 & 75.6±0.1 & \underline{-7.8±0.1} & 53.4±0.3 & -12.5±0.2 & \underline{96.9±0.1} & \underline{-0.7±0.1} & \underline{71.7±0.1} & \underline{-5.8±0.1} \\
\hline
\multirow{1}{*}{Arc}
& TPP & \textbf{92.7±0.1} & \textbf{0.0±0.1} & \textbf{76.5±0.1} & \textbf{0.0±0.1} & \textbf{85.4±0.1} & \textbf{0.0±0.0} & \textbf{99.4±0.1} & \textbf{0.0±0.1} & \textbf{93.0±0.1} & \textbf{0.0±0.1} \\
\hline
\end{tabular}}
\end{table}

\section{Experiments and Analysis}

Most existing research reports impressive performance in the task-incremental setting, while lacking the reported performance in the more challenging class-incremental setting. Furthermore, different methods are likely to employ different datasets, data splitting strategies and some common hyper-parameters, making it difficult to compare their performance via the reported performance. Therefore, we perform a comparison experiment for some typical methods on four benchmark datasets in the task-incremental and class-incremental setting.

\begin{figure}[htp]
\centering
\includegraphics[width=0.7 \textwidth]{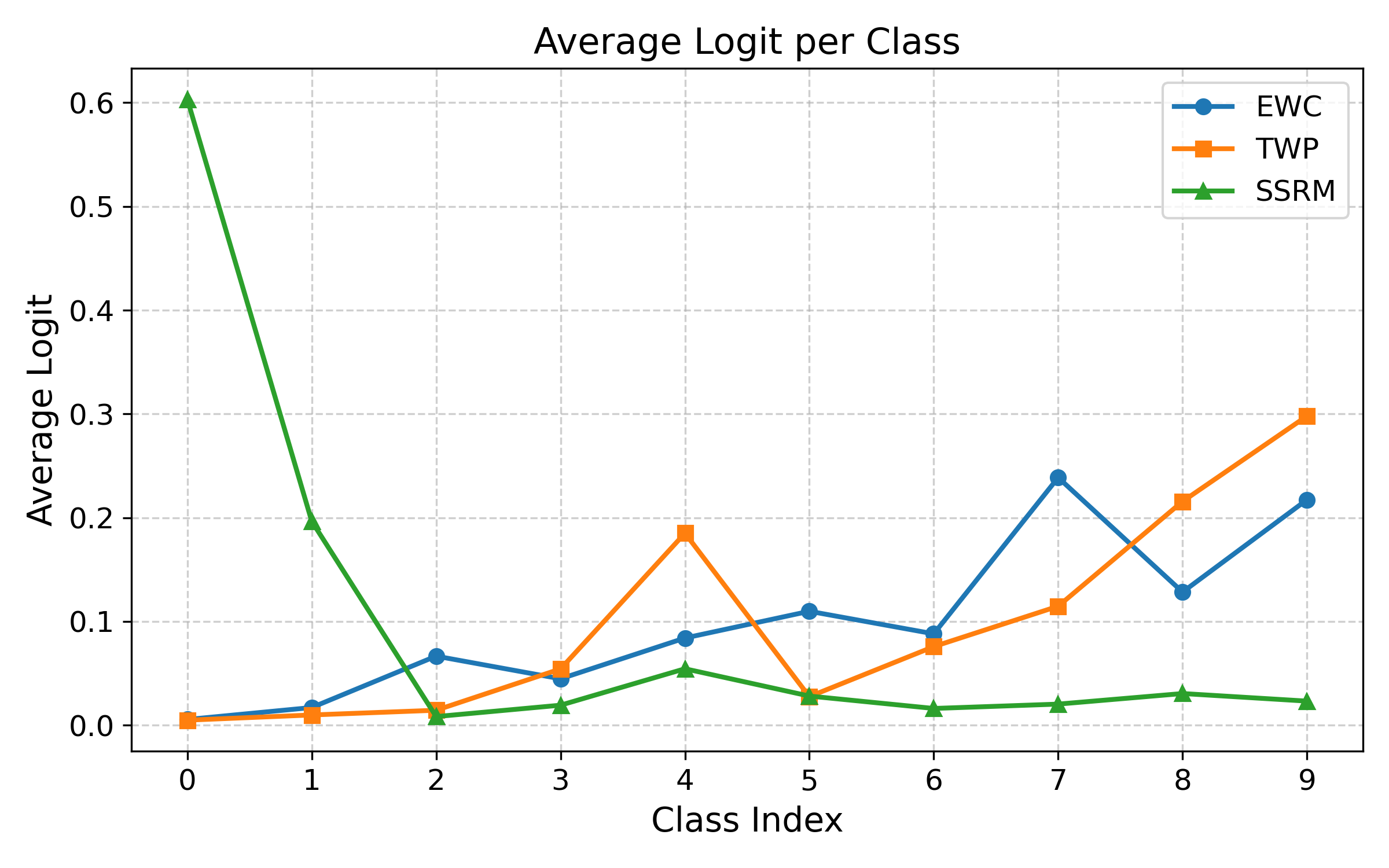}

\caption{Logits visualization of the last layer units of EWC,TWP, and SSRM in the testing process of the first tasks after learning 5 tasks on Corafull dataset under the Class-IL setting.}
\label{logit}
\end{figure}

\subsection{Experimental Setting} 

\begin{table}[htp]
\centering
\caption{Statistics of datasets.}
\label{data_stat}
\small
\begin{tabular}{cccccc}
\hline
Datasets & Nodes     & Edges       & Features & Classes & Tasks \\ \hline
Corafull & 19,793    & 130,622     & 8,710    & 70      & 35    \\
Citeseer & 3,327     & 9,928       & 3,703    & 6       & 3     \\
OGB-Arxiv    & 169,343   & 1,166,243   & 128      & 40      & 20    \\
Reddit   & 227,853   & 114,615,892 & 602      & 40      & 20    \\
OGB-Products & 2,449,028 & 61,859,036  & 100      & 46      & 23    \\ \hline
\end{tabular}
\end{table}

\subsubsection{Datasets and Preprocessing}
Following previous works \cite{CGLB, SSM, puma}, we evaluate these representative methods on five node classification datasets, with detailed statistics summarized in Table \ref{data_stat}. Each dataset is split into a sequence of tasks, where each task contains two unique classes. For each task, 60\% of the data is used for training, 20\% for validation, and 20\% for testing. There are no inter-task edges connecting nodes between different tasks. In this study We adopt a transductive learning setting, where the testing nodes are visible during training but their labels remain hidden.

\subsubsection{Implementation Details}
In this study, we use 2-layer GCN and SGC backbones with 256 hidden units for all CL methods to ensure a fair comparison. all baselines are implemented using the CGLB benchmark \cite{CGLB}, with PyTorch and the Deep Graph Library (DGL) \cite{dgl} as the core Python libraries. All models are trained for 200 epochs without batch processing, i.e., the entire graph is processed once per iteration. The GNN backbones are uniformly configured with 2 layers and a hidden dimension of 256. The learning rate is set to 0.005, the weight decay to 0.0005, and the Adam optimizer is employed.

For replay-based methods (ERGNN, SSM, CaT, PUMA, PDGNN), the memory budget is set to 50 nodes per class for the Corafull and Citeseer datasets, and 100 nodes per class for the Arxiv, Reddit, and Products datasets. For EWC, the regularization strength is searched over the range [100,1000,10000]. For TWP, the two regularization terms corresponding to the loss function and the topological information are each searched over [100,1000,10000]. For SSM, which selects ego-subgraphs centered around a list of core nodes, assigning the same number of center nodes across methods may cause unfairness. To address this, the memory bank is partitioned as 10\%, 20\%, and 20\% for center nodes, first-hop neighbors, and second-hop neighbors, respectively. For CaT, the number of encoders is set to 1000, and the learning rate for pseudo-subgraph generation is 0.001. For PUMA, the number of encoders is set to 100, with the same learning rate for the pseudo-subgraph. In each iteration, the pseudo-subgraph is updated 10 times, following the configuration in the original paper \cite{puma}. For TPP, we adopt the optimal hyperparameters reported in the original paper: the number of prompts is set to 3, the edge drop rate to 0.2, and the feature drop rate to 0.3. All experiments are conducted five times, and the mean value and standard deviation are reported. All experiments are performed on a NVIDIA A40 GPU.

\subsection{Main Results}
\label{analysis}

The empirical results across four benchmark settings, Task-IL and Class-IL with both GCN and SGC backbones, highlight several key patterns and limitations in current continual graph learning methods. First, regularization-based approaches (e.g., EWC, TWP, SSRM) suffer severe degradation under the Class-IL setting. While they perform moderately well in Task-IL with acceptable accuracy and forgetting (e.g., SSRM achieves 91.7\% AP with 0.4\% AF on CoraFull), their performance collapses in Class-IL, often dropping below 10\% AP with extreme forgetting (e.g., EWC reaches only 4.9\% AP and 90\% AF on Arxiv). Figure \ref{logit} visualizes the output logits of EWC, TWP, and SSRM on the first task after learning five tasks. To correctly classify nodes from the first task, the logits for classes 0 and 1 should be higher than others. However, the logits of EWC and TWP are nearly zero, overshadowed by the logits of later classes. In contrast, SSRM retains appropriately high logits for the correct classes. A possible reason is that EWC and TWP apply regularization across all parameters, including those in the output layer. Their design assumes a shared output head across tasks, which is violated in Class-IL. In comparison, SSRM imposes regularization before the second hidden layer, allowing the output layer to flexibly adapt to new knowledge. The architecture-based method TPP demonstrates consistently superiority in Class-IL, achieving both the highest AP and zero forgetting across all datasets and backbones (e.g., 93.1\% AP on CoraFull and 99.4\% AP on Reddit). This success stems from its prompt-based architecture, which isolates task-specific subspaces and enables accurate task inference during testing, thereby eliminating inter-task interference.

Replay-based methods, such as CaT, PDGNN, and PUMA, strike a better balance between stability and plasticity than regularization, performing competitively in both settings. However, differences emerge within this family. For example, although PUMA enhances CaT by adding pseudo-samples, it surprisingly underperforms CaT on several Class-IL benchmarks. This may be attributed to unreliable class boundaries in early incremental stages, in Class-IL, pseudo-samples risk introducing noisy gradients when task-level confidence is low. Another notable case is ERGNN-MF, which shows significantly lower performance than its variants. This is likely due to its memory selection strategy, which prioritizes nodes near class feature centroids, resulting in insufficient diversity and ineffective knowledge retention.

Additionally, memory limitations restrict the scalability of several methods. On large datasets like Reddit and Products, entries for TWP, SSRM, and PUMA are missing due to out-of-memory (OOM) errors. This can be explained by their computational requirements: SSRM’s MMD calculation grows quadratically with sample size (Although CaT and PUMA involve MMD computation either, they use a memory-friendly linear kernel). TWP needs to preserve fine-grained edge interactions across tasks, leading to excessive memory overhead.

\begin{figure}[htp]
\centering
\includegraphics[width= \textwidth]{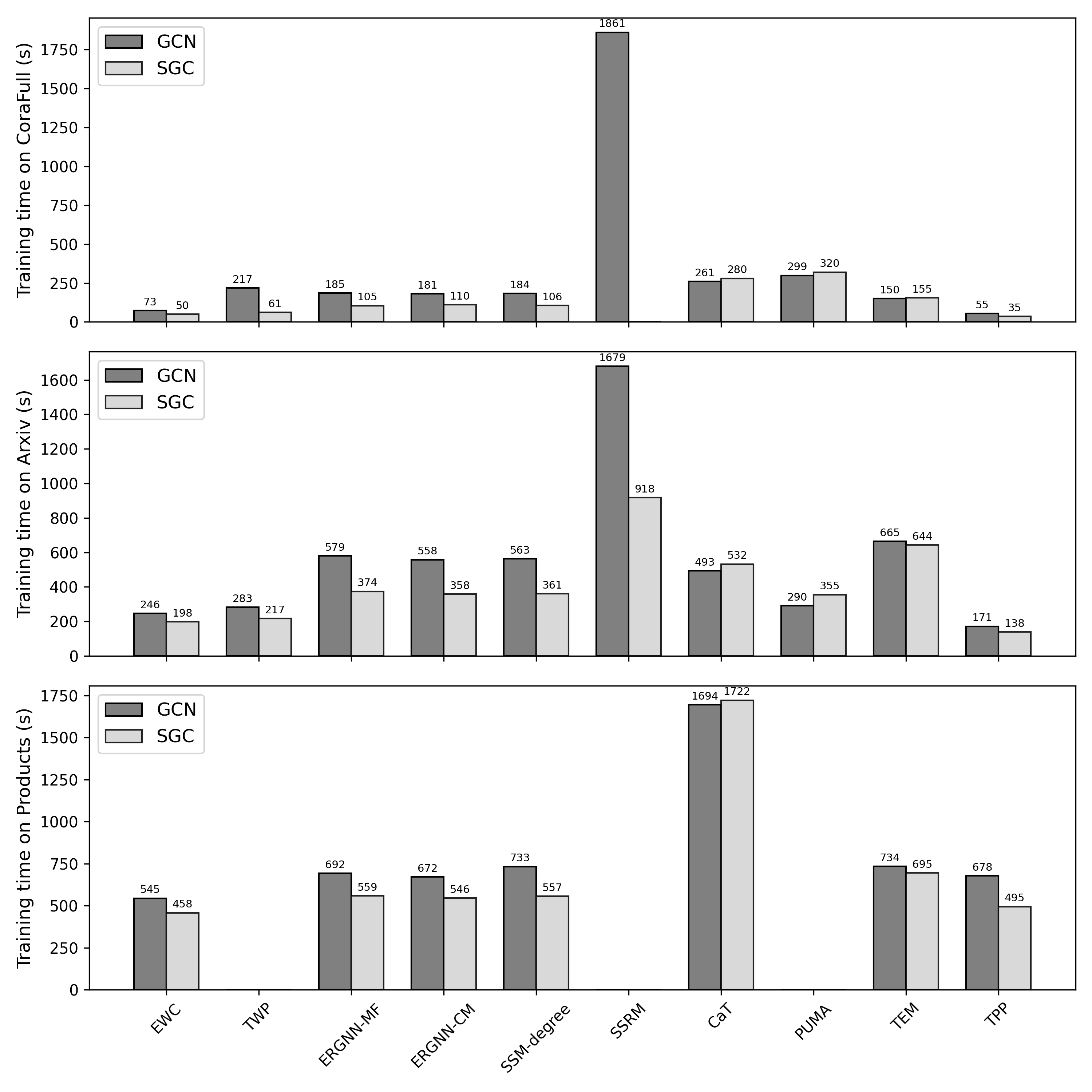}
\label{logits}
\caption{Comparison of training time (s) of methods on three datasets under the Task-IL setting.}
\label{training_time}
\end{figure}

\subsection{Time Efficiency}

Figure~\ref{training_time} presents the training time of various continual graph learning methods across three benchmark datasets. A consistent ranking emerges across all settings: regularization-based and prompt-based approaches are generally the most efficient, while replay-based methods, particularly those incorporating graph condensation or large memory buffers, incuring significantly higher computational costs. For instance, both TPP (task profiling and prompting) and EWC complete training in under 100 seconds on CoraFull, making them among the most efficient. In contrast, SSRM requires substantially more time (e.g., $\approx$ 1861s on CoraFull), primarily due to its reliance on maximum mean discrepancy (MMD) computation. Given two input matrices of size $[n, d]$ and $[m, d]$, the MMD implementation has a time complexity of $O(d(n^2 + mn + m^2))$, which grows quadratically with the number of nodes in both the incoming graph and the memory buffer.

CaT is another clear outlier in terms of cost. Its graph condensation module repeatedly transforms input graphs into embeddings, introducing substantial overhead from repeated message passing. These trends persist across all datasets (CoraFull, ArXiv, and Products): training time scales with dataset size, but the relative efficiency ranking remains stable. Notably, TPP consistently remains the fastest, as it avoids memory replay and trains compact prompts independently. In contrast, SSRM and CaT remain the most time-consuming, especially on large graphs like Products.

It is expected that generative replay-based methods (e.g., CaT, PUMA, DeLoMe, SGNN-GR) incur higher runtime due to the overhead of training generative models. It is acceptable if they can achieve superior performance than others. However, as shown in Tables~\ref{result_3} and \ref{result_4}, TPP achieves superior performance with much lower training cost, highlighting a promising trade-off between efficiency and effectiveness.

\section{Open Issues of CGL}

% \subsection{Identify the Research Gap}

% Once the setting is clear, the research gap should be identified—whether in performance, efficiency, or application scope. In terms of performance, SSM~\cite{SSM} notes that ER-GNN stores isolated nodes, limiting structural preservation. It improves this by sampling central nodes with their neighbors to enhance connectivity. SEM~\cite{SEM} builds on this by using Ricci curvature to avoid sampling irrelevant neighbors. CaT~\cite{CaT} highlights the inability of replay methods to retain global structure and instead condenses graphs into compact, informative subgraphs. For efficiency, PUMA~\cite{puma} replaces CaT's GNN encoder with a Laplacian-smoothed MLP, cutting training time. PDGNN~\cite{PDGNN} compresses ego-graphs into fixed-size temporal embeddings, reducing memory usage. In terms of scope, TrafficStream~\cite{TrafficStream} pioneers the application of CGL in traffic flow prediction, expanding CGL beyond academic benchmarks.

\label{CGL_issue}

We have investigated mainstream CGL approaches and provided a comprehensive discussion of CGL from a technical perspective. In this section, we summarize some open issues and challenges in CGL. 

\subsection{Over-stability problem}
Previously most methods seek to overcome the catastrophic forgetting problem since this directly affects the performance of CGL. Plasticity is sometimes underestimated because warm-start can achieve acceptable performance. Currently since strategies to maintain model's stability have been widely studied, research attention on enhance model's plasticity have increasingly raised. 

\subsection{Efficient prototype $\&$ generative replay-based approach}
As shown in Figure \ref{timeline}, replay-based methods have dominated recent years due to their plasticity and ease of implementation. However, experience replay faces privacy concerns and struggles to fully retain past knowledge. To address this, recent methods employ prototype generation or pseudo-data synthesis. While generative replay improves accuracy, it often incurs high computational costs. For instance, PUMA \cite{puma} takes longer than joint training on Arxiv. Thus, developing efficient prototype- or generation-based replay methods is a promising future direction.

\subsection{Prompt-based CGL}
Prompt-based continual learning (CL) methods, such as \cite{wang2022learning}, utilize a frozen pre-trained large model to store shared knowledge, while task-specific or domain-specific knowledge is encoded in trainable prompts. Unlike traditional CL approaches like regularization- and replay-based methods, prompt-based CL can automatically select relevant knowledge for each sample without requiring task identifiers, offering greater flexibility. However, a key challenge for prompt-based continual graph learning (CGL) is the lack of pre-trained large models for graphs. While some models, such as \cite{tang2024graphgpt}, have emerged, they remain limited in number.

\section{Conclusion}

This survey provides a comprehensive overview of Continual Graph Learning, analyzing foundational concepts, key approaches, and open challenges. It serves as a resource for both newcomers and experts by offering an in-depth critique of existing methodologies.

Benchmarking across five datasets and two continual learning settings reveals several critical insights. Regularization methods, while efficient, struggle in Class-IL as they lose class distinction with growing label spaces. Replay strategies using graph condensation or generative models offer a better balance of stability and adaptability than raw sampling, though widespread adoption is hindered by high training costs. Architecture-based approaches, such as prompt learning, are efficient and effective at preventing forgetting but currently rely on task-specific designs.

To bridge the gap between current methodologies and future demands, this survey highlights critical avenues for innovation. A primary challenge lies in addressing the stability-plasticity dilemma more holistically. This requires shifting focus from merely reducing forgetting to mitigating over-stability, thereby enhancing adaptability to new structural patterns. Simultaneously, to improve practical scalability, the community is expected to prioritize the development of lightweight generative replay strategies that eliminate the heavy computational burden associated with current leading techniques. Furthermore, the convergence of prompt-based methods with pre-trained Large Graph Models (LGMs) represents a transformative, rehearsal-free direction, particularly as universal pre-training frameworks for graphs continue to evolve.

Finally, we acknowledge certain limitations. The survey prioritizes regularization and replay techniques, giving less attention to architecture-based methods, and relies on a subset of experimental comparisons. Future work should address these gaps by expanding the scope of evaluation and incorporating the latest architecture-based techniques to further advance CGL.

\section*{Declaration of Competing Interest}
The authors declare they have no known competing financial interests or personal relationships that could have appeared to influence the work reported in this paper.

\section*{Acknowledgements}
This research is supported by the Xi’an Jiaotong Liverpool University (XJTLU) PGRS PhD Scholarship Fund (No.PGRS2006022).
%% Loading bibliography style file

% \bibliographystyle{elsarticle-num}

% % \bibliographystyle{cas-model2-names}
% \bibliography{ref}

% % Loading bibliography database

\end{document}